\documentclass[lettersize,journal]{IEEEtran}
\usepackage{amsmath,amsfonts}
\usepackage{algorithmic}
\usepackage{algorithm}
\usepackage{array}
\usepackage[caption=false,font=normalsize,labelfont=sf,textfont=sf]{subfig}
\usepackage{textcomp}
\usepackage{stfloats}
\usepackage{url}
\usepackage{verbatim}
\usepackage{graphicx}
\usepackage{cite}

\usepackage{amssymb} 
\usepackage{color}
\usepackage[table]{xcolor}
\usepackage{enumitem}
\usepackage{bm}
\usepackage{hyperref}

\usepackage{booktabs}     
\usepackage{multirow}     
\usepackage{pifont} 
\usepackage{colortbl}
\newcommand{\cmark}{\ding{51}}
\definecolor{darkgreen}{rgb}{0.0, 0.5, 0.0}

\usepackage{tabularx}

\begin{document}

\title{Polycepta: Object-Centric Appearance Estimation for Multi-Object Tracking}

\author{Mohamed Nagy,~\IEEEmembership{Member,~IEEE,} Naoufel Werghi,~\IEEEmembership{Senior Member,~IEEE,} Jorge Dias,~\IEEEmembership{Senior Member,~IEEE,} Majid Khonji,~\IEEEmembership{Member,~IEEE}
\thanks{Manuscript received ...This research was supported by the Center for Autonomous Robotic Systems, Khalifa University of Science and Technology. KU-CARS}
\thanks{The authors are with Khalifa University, Abu Dhabi, UAE, (e-mail: mohamed.nagy@ieee.org; naoufel.werghi@ku.ac.ae; jorge.dias@ku.ac.ae; majid.khonji@ku.ac.ae).}
}

\markboth{Submitted manuscript, 2026}
{Shell \MakeLowercase{\textit{et al.}}: A Sample Article Using IEEEtran.cls for IEEE Journals}


\maketitle

\begin{abstract}
The tracking-by-detection paradigm in multi-object tracking (MOT) typically relies on static appearance descriptors to complement motion estimation. However, these descriptors are frame-independent, limiting their robustness as visual cues. Since such descriptors are often obtained from computationally intensive pretrained backbones, real-time MOT systems frequently abandon appearance cues altogether and rely solely on motion prediction and geometric association. In this work, we introduce Polycepta, an object-centric appearance state estimation framework that reformulates appearance modeling as a recursive estimation problem rather than a frame-wise matching task. Polycepta constructs and continuously updates an independent appearance state for each tracked object, enabling future appearance representations to be estimated from accumulated observations. Polycepta is encouraged to learn the appearance-state construction of object-specific representations rather than memorize them through a proposed learning strategy, enabling appearance estimation for unseen classes. A key property of Polycepta is that the quality of appearance estimation improves as object states evolve during inference. While conventional appearance descriptors remain static or degrade over time, Polycepta progressively refines appearance estimates as additional observations are accumulated. Extensive experiments on KITTI, the Waymo Open Dataset, and MOT17 demonstrate consistent reductions in identity switches and improvements in tracking performance when integrated into the tracking-by-detection pipelines. Polycepta operates at 90.57 Hz and delivers state-of-the-art performance on the KITTI benchmark, achieving a MOTA of 92.27\%.
\end{abstract}


\section{Introduction}
\IEEEPARstart{M}{ulti}-object tracking (MOT) within the tracking-by-detection paradigm has traditionally relied on state estimation and motion prediction using the Kalman filter (KF), frequently omitting complex visual cues to minimize computational overhead. While modern frameworks~\cite{hybridsort,bytetrack,afmtrack,strongsort,tracktrack} attempt to integrate appearance modeling alongside KF motion modeling, they typically employ static feature extractors—such as Re-identification (ReID) networks \cite{reid}—pre-trained on massive datasets. Although effective, these approaches often introduce substantial computational overhead, which can limit deployment in latency-constrained applications, such as self-driving vehicles, due to their reliance on high-capacity, computationally intensive backbones (e.g., ResNet-50). Consequently, real-time online trackers~\cite{robmot,deepfusion,mctrack} often omit appearance cues entirely, relying solely on trajectory data association to maintain low latency and high operational throughput.
\begin{figure}
    \centering
    \includegraphics[width=\linewidth]{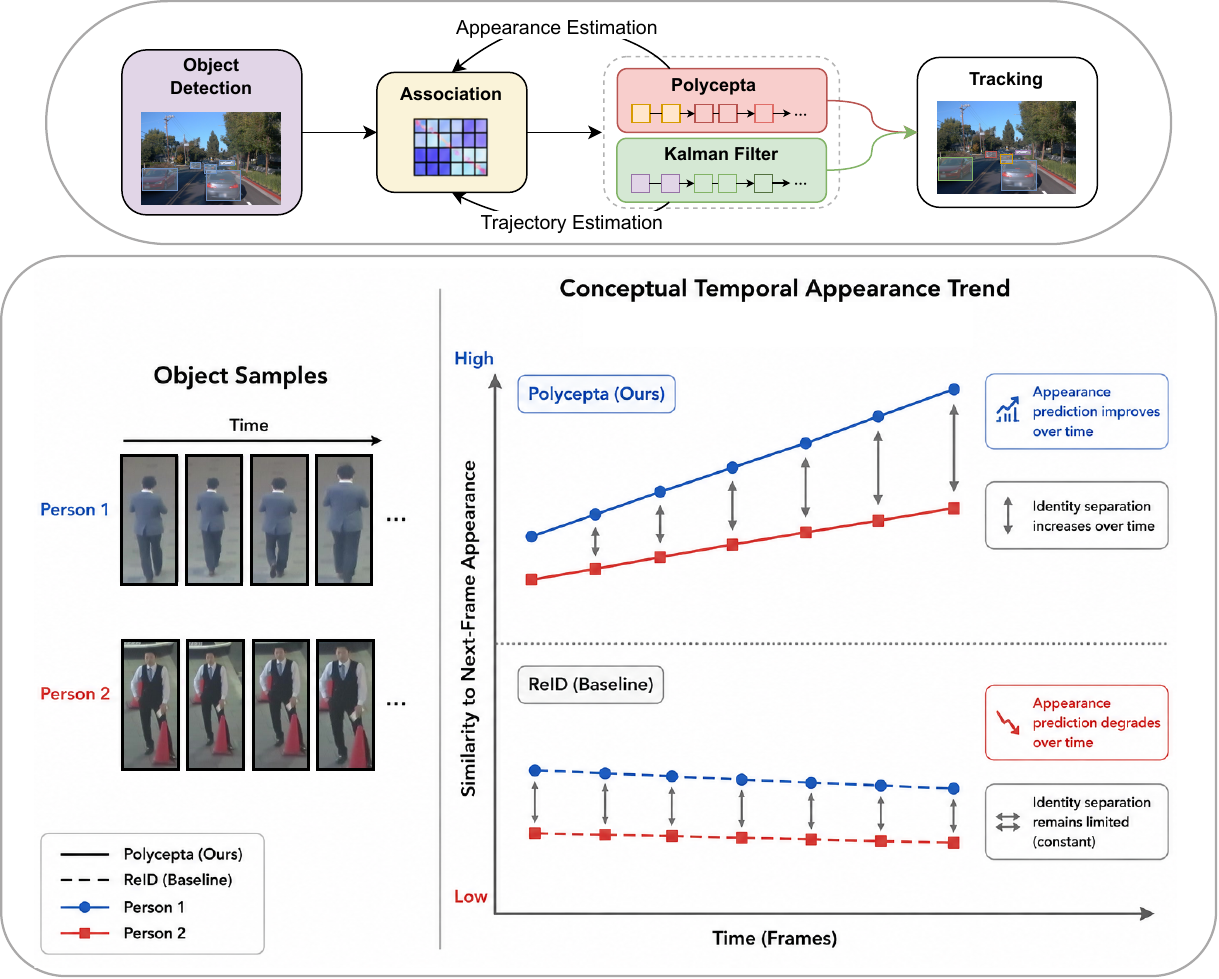}
    \caption{Visual illustration of visual estimation concept of Polycepta, and its integration into the MOT pipelines along with the Kalman filter.}
    \label{fig:integration}
    \vspace{-0.3cm}
\end{figure}
Furthermore, the core limitation of conventional ReID-based approaches within the tracking-by-detection paradigm lies in their rigid frame-by-frame feature extraction and matching. This formulation implicitly assumes that an object's appearance remains static across time—an assumption that directly contradicts the continuous visual evolution that objects undergo in dynamic environments. Consequently, these approaches are brittle under partial occlusions, perspective changes, or environmental noise, all of which distort the extracted feature embeddings (Figure~\ref{fig:integration}).

Conceptually, this stands in stark contrast to motion modeling; the enduring success of the KF lies precisely in its ability to recursively refine an object's kinematic state over time, rather than relying solely on isolated, instantaneous observations. Current appearance cues lack an equivalent recursive estimation mechanism, leaving a critical gap in maintaining temporal visual identity.

To address these limitations, we introduce Polycepta, an object-centric appearance state estimation framework that reformulates appearance modeling as a recursive estimation problem. Instead of associating objects using frame-wise appearance descriptors alone, Polycepta constructs and continuously updates an independent appearance state for each tracked object. These states accumulate appearance information over time and are used to estimate future appearance representations for data association (Figure~\ref{fig:integration}).

To discourage memorization of object-specific appearance histories, all appearance states are reset at the end of each training epoch. This state-erasure strategy forces the model to repeatedly reconstruct appearance states from observations rather than relying on persistent training-time memories. As a result, Polycepta learns state-construction dynamics that transfer across object categories and support appearance estimation for previously unseen classes. 

More broadly, Polycepta extends the state-estimation principle that has long underpinned motion modeling in MOT to the appearance domain. While existing tracking-by-detection frameworks estimate motion recursively and treat appearance as a sequence of independent observations, Polycepta models appearance as a recursively evolving state that accumulates information over time.

The contribution of this work is  summarized as follows:

\begin{enumerate}

\item \textbf{Object-Centric Appearance-State Estimation for MOT:}
We reformulate appearance modeling in the tracking-by-detection paradigm as a recursive state-estimation problem. Unlike conventional ReID-based approaches that treat appearance as a sequence of independent frame-wise observations, Polycepta constructs, maintains, and continuously updates an appearance state for each tracked object, enabling appearance information to accumulate over time and supporting future appearance estimation for data association.

\item \textbf{Polycepta Framework:}
We introduce Polycepta, an appearance-state estimation framework that combines structured state-space dynamics, relational appearance correlation in the Fourier domain, and adaptive appearance-state updates to model appearance evolution while maintaining real-time operation. The framework is designed to complement motion-centric tracking-by-detection systems without modifying their underlying motion-estimation pipeline (Figure~\ref{fig:integration}).

\item \textbf{State-Erasure Learning:}
We introduce a state-erasure learning strategy that resets appearance states during training to discourage memorization of object-specific appearance histories. This encourages the learning of appearance-state construction mechanisms and enables robust generalization to previously unseen object categories.

\item \textbf{Temporal Appearance Refinement and Generalization:}
We show that appearance estimation quality improves during inference as appearance states evolve, in contrast to conventional ReID descriptors whose quality remains fixed or gradually degrades over time. Extensive experiments on KITTI, MOT17, and the Waymo Open Dataset (WOD) demonstrate consistent reductions in identity switches and improvements in tracking performance across datasets, detector configurations, and previously unseen object classes while maintaining real-time operation at 90.57 Hz.

\end{enumerate}

\section{Related Work}
\label{sec:related_work}

\subsection{Tracking-by-Detection Paradigm}
MOT frameworks~\cite{hybridsort,bytetrack,afmtrack,strongsort,10777493,peng2024pnasmot} have leaned heavily on feature extraction ReID to maintain track continuity across temporal gaps. These methods typically employ static, high-capacity feature extractors to generate visual embeddings for frame-to-frame matching. They ultimately treat appearance as an independent observation that serves as a visual cue for inferring associations, thereby overlooking temporal changes in objects' appearance across consecutive frames, which limits the effectiveness of these methods' visual cues. 

On the other hand, to satisfy the low-latency constraints of real-time autonomous systems, 2D and 3D online trackers~\cite{robmot,fasttrack,mctrack,kalman1960,bytetrack,nagy} frequently prioritize motion-centric paradigms based on linear state estimators, such as the KF, thereby entirely forgoing visual cues. While hybrid architectures~\cite{10771607} aim to fuse appearance cues with motion prediction via end-to-end learning, the computational overhead often limits their adoption in real-time tracking systems. Yet, the appearance cues in these methods are treated as static descriptors. 
\subsection{Tracking-by-Propagation Paradigm}

Beyond the conventional tracking-by-detection paradigm, recent work has explored tracking-by-propagation frameworks that maintain object identities through trajectory propagation and prediction~\cite{MeMOT,MeMOTR,Samba}. These approaches leverage visual and motion cues to propagate object tracks over time, reducing the reliance on explicit frame-wise data association.

Transformer-based methods such as MeMOT and MeMOTR~\cite{MeMOT,MeMOTR} employ memory propagation through object-level attention mechanisms to maintain track continuity. More recently, Samba~\cite{Samba} incorporates state-space modeling based on Mamba~\cite{gu2023mamba} to improve computational efficiency while continuing to propagate object trajectories through time. 

Despite recent advances in tracking-by-propagation, tracking-by-detection remains a dominant paradigm for real-time MOT due to its favorable accuracy-efficiency tradeoff. Consequently, this work focuses on improving appearance modeling within the tracking-by-detection paradigm, where strong accuracy-efficiency tradeoffs remain particularly important for real-time MOT.

Polycepta addresses these limitations by reformulating appearance modeling as a recursive state-estimation problem within the tracking-by-detection paradigm. Rather than representing appearance as a sequence of independent observations, Polycepta constructs and recursively updates an object-centric appearance state for each tracked object. These appearance states accumulate information over time and are used to estimate future appearance representations for data association. As a result, Polycepta provides lightweight appearance cues that can be integrated into existing motion-centric tracking-by-detection frameworks without modifying their underlying motion-estimation pipeline.
\section{Method}
\begin{figure*}[!t]
\centering
\includegraphics[width=\textwidth]{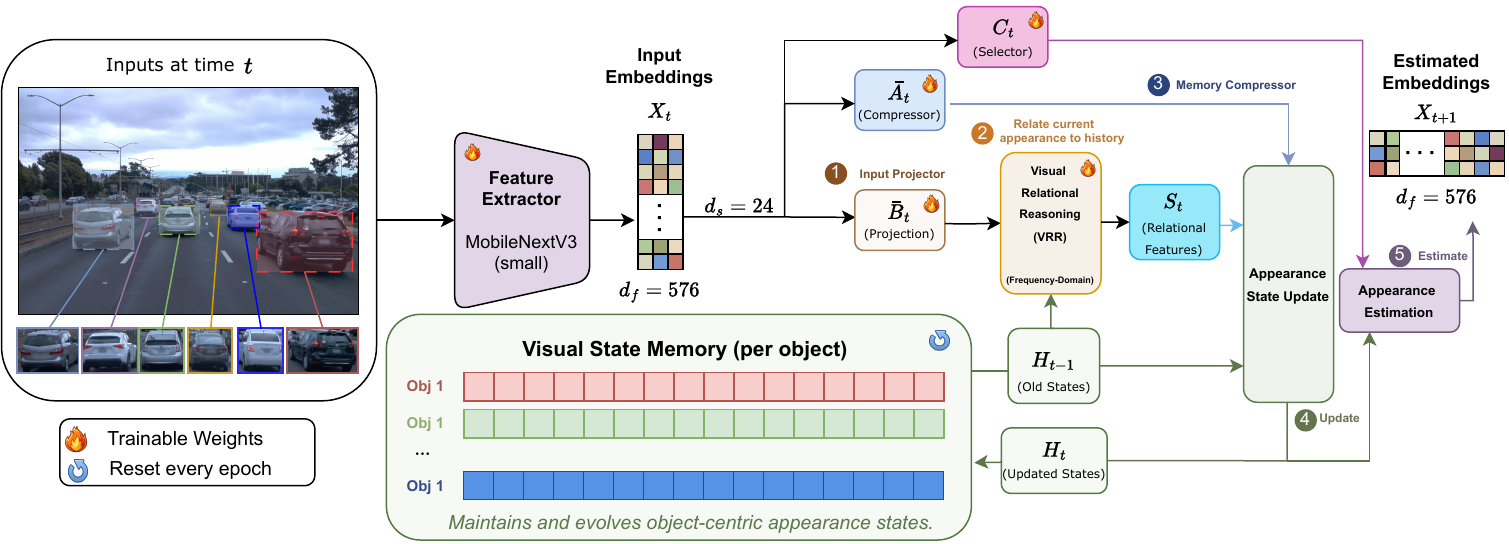}
\caption{A high-level overview of the Polycepta architecture during inference. Initially, incoming objects are cropped and grouped into a single batch for parallelism. $X_t$ matrix is formed to hold feature embeddings of the objects obtained from a lightweight feature embedding (MobileNetV3-small). The feature embeddings are projected into a lower dimension $B_t$ (state space $d = 24$). Visual relational reasoning (VRR) correlates with the observed feature embeddings and their historical appearance memory (state) $H_{t-1}$ in a signal frequency domain to obtain generated synthetic features $S_t$. Next, the appearance states are updated in the appearance state update module. Lastly, the updated appearance states are used to obtain the next appearance estimation $\hat{X}_t$. }
\label{fig:polycepta}
\vspace{-0.3cm}
\end{figure*}
\subsection{Polycepta: Architecture Overview}
The overall architecture of Polycepta comprises three interconnected modules: Visual Relational Reasoning (VRR), Appearance State Update, and Appearance Estimation, as illustrated in Figure~\ref{fig:polycepta}.  The framework is designed to recursively construct and update an appearance state for each tracked object, enabling future appearance representations to be estimated from accumulated observations.

At each discrete time step $t$, the system receives a batch of cropped image bounding boxes corresponding to $n$ detected objects. This batch is processed through a highly efficient, low-latency backbone, specifically \textit{MobileNetV3\_small}~\cite{mobilenet}, to extract an instantaneous feature embedding matrix $X_t=\{x_t^i\}^n_{i=1} \in \mathbb{R}^{n \times 576}$, where $x_t^i$ represents the feature embeddings of the $i$-th object.

Next, the object-centric feature embeddings are projected from the high-dimensional feature space $d_f=576$ into a state space of dimension $d_s=24$ through the linear projector $B_t$.  The collective appearance states of all active tracking targets are maintained in an appearance state matrix $H_t = \{h_t^i\}_{i=1}^n$, where $h_t^i \in \mathbb{R}^{d_s}$ denotes the recursively updated appearance state of the $i$-th object. A time-dependent step size parameter $\Delta_t$ is utilized to discretize the continuous-time matrices $A_t$ and $B_t$ into their discrete-time counterparts, $\bar{A}_t$ and $\bar{B}_t$, following standard state-space discretization protocols~\cite{gu2020hippo}, as detailed in Section~\ref{subsec:memory_compression}.

The operational cycle of Polycepta proceeds sequentially through three distinct architectural stages:
\begin{enumerate}
    \item \textbf{Visual Relational Reasoning (VRR):} The VRR module maps the current discretized appearance features $\bar{B}_t$ and the historical appearance states $H_{t-1}$ into a learnable Fourier signal space. It performs frequency-domain correlation between recent observations and historical appearance states for each object (in-parallel), and generates relational features $S_t = \{s_t^i\}_{i=1}^n$ (Section~\ref{subsec:vssp}).
    \item \textbf{Appearance State Update:} Taking the current discretized appearance features $\bar{B}_t$, the prior history $H_{t-1}$, and the relational features $S_t$ as inputs, this module applies an adaptive gated mechanism to compute the updated object-centric appearance states $H_t$ (Section~\ref{subsec:ocss}).
    \item \textbf{Appearance Estimation:} Finally, the refined appearance states $H_t$ are passed to the estimation block to infer the anticipated next appearance descriptors for the subsequent time, yielding $\hat{X}_{t+1} \in \mathbb{R}^{n \times 576}$ (Section~\ref{subsec:appearance_estimation}).
\end{enumerate}
Collectively, this architecture allows Polycepta to recursively maintain an appearance state for each tracked object and estimate future appearance representations from accumulated observations. In contrast to frame-wise appearance descriptors, the resulting appearance states evolve throughout inference, progressively incorporating information gathered over time.

\section{Projection and Discretization}
\label{subsec:memory_compression}
To map instantaneous visual features into a temporal state space, the raw feature embeddings $X_t$ are passed through a normalization layer. As formalized in Equation~\ref{eq:32_proj}, the normalized representation $z_t$ is projected via learnable parameter matrix $W_B$ to obtain $B_t$, the time-dependent discretization step size $\Delta_t$, and the selection vector $C_t$ used for downstream appearance estimation:
\begin{equation}
\begin{aligned}
z_t &= \text{LayerNorm}(X_t) \\
\Delta_t &= \text{softplus}(W_{\delta} z_t + b_{\delta}) \\
B_t &= \text{LayerNorm}(W_B z_t + b_B) \\
C_t &= W_C z_t + b_C
\end{aligned}
\label{eq:32_proj}
\end{equation}
where $W_{\delta}, W_B, W_C$ and $b_{\delta}, b_B, b_C$ denote the learnable weights and biases of their respective projection layers.

The appearance state evolves according to a state transition matrix $A_t$ and an input projection matrix $B_t$. The resulting state dynamics govern how appearance information is accumulated and updated over time. Following prior state-space formulations~\cite{gu2020hippo,gu2023mamba}, we construct $A_t$ by combining a structured HiPPO transition matrix with a learnable diagonal adaptation. The HiPPO component provides a fixed state-space structure, while the learnable diagonal enables the transition dynamics to adapt during training.

Specifically, let $\theta \in \mathbb{R}^{d_s}$ parameterize the learnable diagonal component and let $A_{\text{hippo}} \in \mathbb{R}^{d_s \times d_s}$ denote the HiPPO transition matrix. The resulting state transition matrix is defined as
\begin{equation}
\begin{aligned}
A_{\text{diag}} &= \exp(\theta), \\
A_t &= A_{\text{hippo}} - \text{diag}(A_{\text{diag}}).
\end{aligned}
\end{equation}

The continuous-time matrices are subsequently discretized using the learned step size $\Delta_t$:
\begin{equation}
\begin{aligned}
\bar{A}_t &= I + \Delta_t A_t, \\
\bar{B}_t &= \Delta_t B_t.
\end{aligned}
\end{equation}
\section{Visual Relational Reasoning}
\label{subsec:vssp}

\begin{figure}
    \centering
    \includegraphics[width=0.7\linewidth]{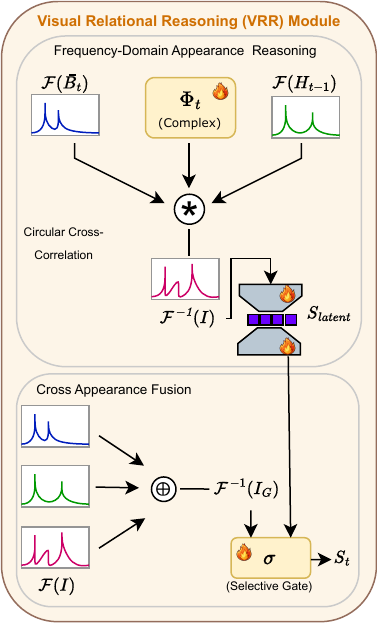}
    \caption{Overview of the VRR architecture. Initially, the recent appearance embeddings $\bar{B}t$ and their historical visual-semantic embeddings $H{t-1}$ are transformed using the Fourier transform into a signal space, and a circular cross-correlation is applied to match similarities between the signals to obtain a relational signal $F(I_t)$, compressed into a smaller latent space $d_s=6$. Lastly, the synthetic features $S_t$ are generated through a selective signal gate $I_G$.}
    \label{fig:vrr}
    \vspace{-0.4cm}
\end{figure}

Updating appearance states using only the current observation $\bar{B}_t$ may fail to fully exploit information accumulated in previous appearance states. To address this, the VRR module explicitly correlates the current observation with each object's historical appearance state before state updates are performed, demonstrated in Figure~\ref{fig:vrr}. The resulting relational features summarize the agreement between recent observations and accumulated appearance information, and are subsequently used to refine the evolution of appearance states.

\subsection{Frequency-Domain Appearance Reasoning}
The VRR module models the correlation between the discretized observed appearances $\bar{B}_t \in \mathbb{R}^{n \times d_s}$ and the latest object appearance states $H_{t-1} \in \mathbb{R}^{n \times d_s}$ to generate synthetic features $S_t \in \mathbb{R}^{n \times d_s}$. While a naive construction of $d_s \times d_s$ correlation matrices between each pair $(\bar{b}_t^i, h_{t-1}^i)$ would incur $O(d_s^2)$ cost per object $i$ and scale poorly with the number of objects $n$, VRR performs this reasoning in the frequency domain. Specifically, both $\bar{B}_t$ and $H_{t-1}$ are transformed using the Fast Fourier Transform (FFT)~\cite{frigo2005design}, and circular cross-correlation is applied, reducing the complexity to $O(d_s \log d_s)$~\cite{stockham1966high}. Importantly, all operations are implemented as batch-parallel matrix operations over all objects, avoiding any object-wise sequential processing.\\
To make the captured correlation context learnable, VRR applies a learnable complex-valued filter $\Phi_t^i$ to each object $i$ in the frequency domain. For clarity, we present the formulation for a single object, while in practice, the operation is vectorized and applied in parallel to all objects:

\begin{equation}
    \Phi_t^i = \theta_{re}^i + i \theta_{im}^i, \quad 
    \theta_{re}^i, \theta_{im}^i \in \mathbb{R}^{\lfloor \frac{d_s}{2} \rfloor + 1}
    \label{eq:33_complex_filter}
\end{equation}

Given the discretized appearance matrix $\bar{B}_t$ and the previous appearance state matrix $H_{t-1}$ for all objects, VRR transforms both into the frequency domain and applies a learnable complex-valued filter $\Phi_t$ to perform circular cross-correlation in a batch-parallel manner. This operation captures interactions between current observations and historical appearance states at the object level, forming the frequency-domain interaction $\mathcal{F}(I_t)$, which is then projected back to the state space as $I_t$:

\begin{equation}
\begin{aligned}
\mathcal{F}(\bar{B}_t) &= \text{FFT}(\bar{B}_t) \\
\mathcal{F}(H_{t-1}) &= \text{FFT}(H_{t-1}) \\
\mathcal{F}(I_t) &= \mathcal{F}(H_{t-1}) \odot \mathcal{F}(\bar{B}_t) \odot \Phi_t \\
I_t &= \text{LayerNorm}\!\left(\text{FFT}^{-1}\!\left(\mathcal{F}(I_t)\right)\right)
\end{aligned}
\end{equation}

\subsection{Cross Appearance Fusion}
The interaction tensor $I_t$ encodes relational cues between recent and historical visual appearances; however, not all relational information contributes equally to appearance-state updates. To selectively retain informative interactions, a fusion gate $G$ is constructed that integrates recent visual features $\mathcal{F}(\bar{B}_t)$, historical appearance-state features $\mathcal{F}(H_{t-1})$, and the relational interaction $\mathcal{F}(I_t)$ in the frequency domain (Cross Appearance Fusion in Figure~\ref{fig:vrr}).

\begin{equation}
\begin{aligned}
G_{\text{fused}} &= \text{FFT}^{-1}\!\left(\mathcal{F}(\bar{B}_t) + \mathcal{F}(I_t) + \mathcal{F}(H_{t-1})\right) \\
G &= \sigma\!\left(\text{LayerNorm}(W_g \cdot G_{\text{fused}})\right)
\end{aligned}
\end{equation}

\subsection{Relational Appearance Feature Generation}

To obtain compact relational (synthetic) features  $S_t$ from the relational interaction $I_t$, we first compress $I_t$ into a lower-dimensional latent space $S_{\text{latent}}$ of dimension $d_l < d_s$, where $d_l = 6$ denotes the latent relational dimension, using a learned projection $W_{in} \in \mathbb{R}^{d_s \times d_l}$ followed by a non-linear activation. The latent representation is then projected back to the original dimensionality using $W_{out} \in \mathbb{R}^{d_l \times d_s}$. Finally, the fused selection gate $G$ modulates the reconstructed features to form the final synthetic appearance features.

\begin{equation}
\begin{aligned}
S_{\text{latent}} &= \tanh\!\left(\text{LayerNorm}(W_{in} I_t + b_{in})\right) \\
S_t &= G \odot \left(W_{out} S_{\text{latent}} + b_{out}\right)
\end{aligned}
\end{equation}

The resulting relational features $S_t$ provide appearance-state update cues derived from the relationship between accumulated appearance information and current observations.
\section{Object-Centric appearance state Update}
\label{subsec:ocss}
Each object $i$ maintains its own appearance state, collected as $H_{t-1} = \{h_{t-1}^i\}_{i=1}^n$. The evolution of these appearance states depends on the state transition matrix $\bar{A}_t$,the current appearance observations $\bar{B}_t$, and the relational appearance features $S_t$.\\
First, the state transition matrix is applied to the previous appearance states:

\begin{equation}
\tilde{H}_t = \bar{A}_t H_{t-1}
\end{equation}

Next, the observed appearances and synthetic features are combined via a linear projection to form fused features:

\begin{equation}
\psi_t = \tanh(\text{LayerNorm}(W_K (\bar{B}_t + S_t)))
\end{equation}

Directly adding $\psi_t$ to $\tilde{H}_t$ (as done in prior work~\cite{gu2023mamba,gu2020hippo}) can limit robustness, especially under false detections or partial occlusions. To adaptively balance incoming and stored information, we introduce a gate $U_t$ computed from the previous appearance states and the fused features:

\begin{equation}
U_t = \sigma(\text{LayerNorm}(W_u [H_{t-1}; \psi_t]))
\end{equation}

$U_t$ controls the contribution of the current observation and the accumulated appearance state during state updates. When limited appearance information has been accumulated (e.g., during early observations), the gate favors the incoming features $\psi_t$; conversely, when the current observation is noisy, partially occluded, or unreliable, the gate places greater emphasis on the accumulated appearance state  $\tilde{H}_t$. This element-wise selection mechanism improves robustness to observation noise while preserving appearance information accumulated over time. Finally, $U_t$ updates the appearance state as:

\begin{equation}
H_t = \text{LayerNorm}(U_t \odot \psi_t + (1 - U_t) \odot \tilde{H}_t)
\end{equation}

where $\odot$ denotes element-wise multiplication.
The resulting state $H_t$ serves as the updated appearance representation of the object and is subsequently used for appearance estimation in the next stage.

\begin{figure}[!t]
\centering
\includegraphics[width=\linewidth]{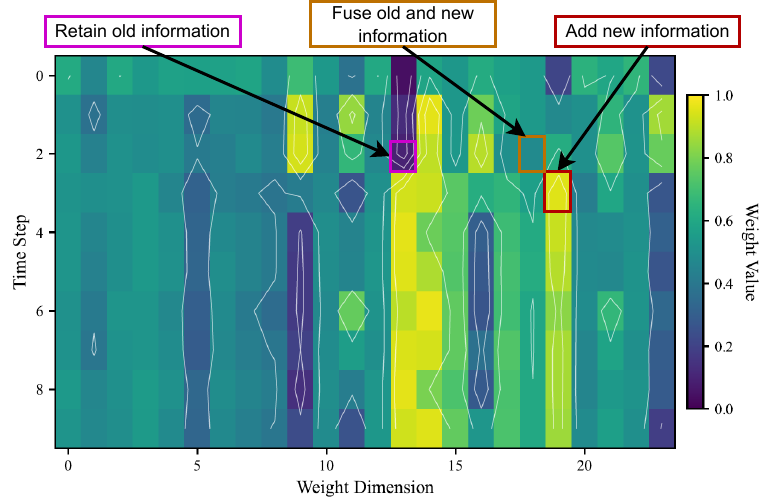}
\caption{Visualization of the gate $U_t$ over time for a single object during the orthogonality ablation study, trained for 5 epochs.}

\label{fig:information_selection}
\end{figure}

Figure~\ref{fig:information_selection} visualizes the evolution of the update gate $U_t$ for a single object over 10 consecutive frames during the orthogonality ablation study. Each row corresponds to the gating pattern at a particular time step. As shown in Figure~\ref{fig:information_selection}, the gate adaptively balances information from the current observation and the accumulated appearance state. Some regions primarily rely on incoming observations (red), while others favor the accumulated appearance state (purple). Intermediate regions exhibit a combination of both sources (orange). 

\section{Appearance Estimation}
\label{subsec:appearance_estimation}
After obtaining the updated object appearance state $H_t = \{h_t^i\}_{i=1}^n$, the selection matrix $C_t$, from Section~\ref{subsec:memory_compression}, is applied to extract appearance-state features $Y_t$ in the state space. The selected features are then projected into the original feature space via $W_{\text{proj}}$ and combined with a learnable residual connection $D_t$ to produce the next appearance estimate $\hat{X}_{t+1}$.
\begin{equation}
    \begin{aligned}
Y_t &= \text{LayerNorm}(C_t \odot H_t) \\
\hat{X}_{t+1} &= (D_t \odot X_t) + (W_{proj} \cdot Y_t)
\end{aligned}
\end{equation}
The estimated appearance $\hat{X}_{t+1}$ is subsequently used as the appearance cue for data association in the next observation step.
\section{Learning Strategy}
\label{subsec:learning_strategy}
\begin{figure*}[!tbh]
\centering
\includegraphics[width=\linewidth]{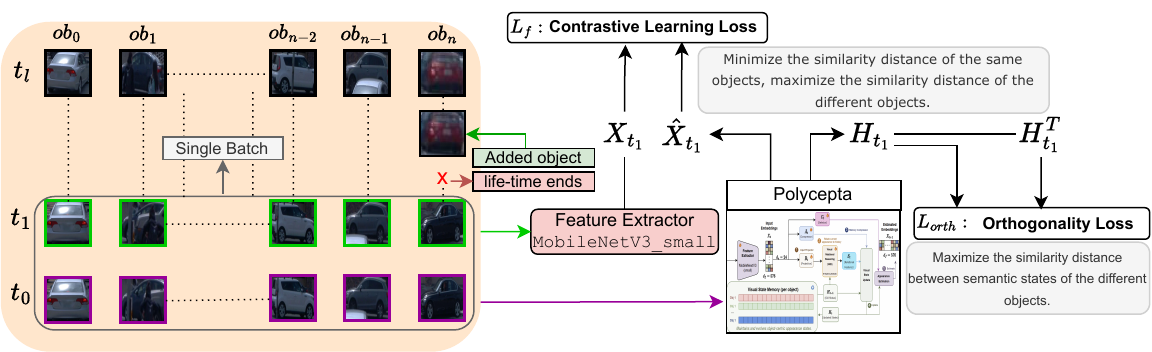}
\caption{ An overview of the training pipeline followed in Polycepta. Objects from the datasets are cropped and sampled temporally. A single batch contains $n$ unique objects. The batch contains the appearance times of the objects and their subsequent appearances (i.e., at $t_0$ and $t_1$). The initial/old appearance \textbf{\textcolor{purple}{(colored purple)}} passes through Polycepta architecture to obtain its estimated appearance embeddings $\hat{X}_{t_1}$ for the subsequent frame. The subsequent appearance  \textbf{\textcolor{darkgreen}{(colored in green)}} passes through MobileNetV3-small to obtain its feature embeddings $X_{t_1}$. The cosine similarity $L_f$ between the subsequent embeddings $X_{t_1}$ and the estimated embeddings $\hat{X}_{t_1}$ is maximized.  Meanwhile, deviating the visual-semantic memories (states) of the objects by minimizing the similarity between each other $L_{orth}$.}
\label{fig:training_pipeline}
\vspace{-0.3cm}
\end{figure*}
\subsection{Dataset Structure and Training Pipeline} 
Polycepta learns object-centric appearance states from temporally ordered appearance observations, which naturally aligns with the sequential nature of online tracking. Accordingly, the training data is organized in an object-level representation, where a sample is a sequence of consecutive appearances of a single object ($ob_i$) over time, as demonstrated in Figure~\ref{fig:training_pipeline}.\\
We construct a dataset from~\cite{geiger2012kitti},~\cite{sun2020waymo},~\cite{mot17}, and~\cite{MOTChallenge20}. Using ground-truth annotations, objects are extracted and cropped frame-by-frame to form temporally ordered appearance sequences, which are stored independently for each object.

In the training pipeline, the dataset is organized as illustrated in Figure~\ref{fig:training_pipeline}, where a set of $n$ objects (batch size) is sorted sequentially, with $n \subseteq \{1,\dots,N\}$ denoting a subset of the total number of objects $N$. Since object lifetimes vary, terminated objects are automatically replaced by selecting from the remaining objects $m \in \{1,\dots,N\} \setminus n$. A sliding window is applied to form training batches, ensuring that each batch contains unique objects. Each batch includes a set of objects $X_{t}$ and their subsequent appearance $X_{t+1}$ (i.e. $t_0$ to $t_1$ as illustrated in Figure~\ref{fig:training_pipeline}). $X_{t+1}$ serves as ground truth for the estimated appearance $\hat{X}_{t+1}$ obtained from $X_{t}$.

Objects $X_t$ (i.e. $X_{t_0}$ in Figure~\ref{fig:training_pipeline}, marked in purple) in the batch pass through the full Polycepta architecture to obtain the estimated appearance $\hat{X}_{t+1}$  and the updated appearance state $H_{t+1}$ (i.e. $\hat{X}_{t_1}$ and $H_{t_1}$ in Figure~\ref{fig:training_pipeline}). 

Their subsequent appearances $X_{t+1}$ (i.e. $X_{t_1}$ Figure~\ref{fig:training_pipeline}, in marked in green) are processed only through the feature extractor to obtain ground-truth embeddings. As described in Section~\ref{subsec:learning_loss_constractive}, a contrastive learning loss is applied between the estimated appearance $\hat{X}_{t+1}$ and the corresponding ground-truth embedding $X_{t+1}$ to pull the estimate toward its true appearance while pushing apart estimates of other objects. Additionally, an orthogonality loss is applied to the appearance states to encourage distinct object-specific state representations and reduce state collapse across objects; an ablation study is conducted in Section~\ref{subsec:ablation_learning_loss}.
\subsection{State-Erasure Learning}

A key objective of Polycepta is to learn to construct object-centric appearance states from observations rather than to memorize object-specific appearance histories. During training, the appearance state of each object evolves recursively as consecutive observations are processed. Consequently, if appearance states are preserved across training epochs, the model may increasingly rely on object-specific information accumulated in the appearance states, reducing the pressure on the learnable parameters to capture the general dynamics of appearance-state construction.

To encourage the learning of appearance-state construction dynamics, we employ a state-erasure strategy. At the beginning of every training epoch, all appearance states are reinitialized to zero:
\begin{equation}
H_0 = \mathbf{0}_{n\times d_s}.
\end{equation}

The training process then proceeds normally, allowing appearance states to evolve recursively as object observations are processed. Once the epoch is completed, all appearance states are discarded and reinitialized before the next epoch begins.

This strategy ensures that the only information retained across epochs resides in the learnable network parameters rather than in the appearance states themselves. As a result, the model is repeatedly forced to reconstruct appearance states from observations and to learn the mechanisms governing appearance-state formation and evolution. Empirically, this strategy improves generalization to previously unseen object categories, suggesting that Polycepta learns appearance-state construction mechanisms that transfer across object classes rather than memorizing category-specific appearance patterns, as demonstrated in Section~\ref{subsec:zero-shot}, where Polycepta maintains tracking performance even when evaluated on classes not observed during training.

\subsection{Contrastive and Orthogonality-Guided Learning}
\label{subsec:learning_loss_constractive}
The training objective is to encourage Polycepta to structure object-centric appearance states to support accurate appearance estimation. To this end, we employ a contrastive feature loss $\mathcal{L}_f$ together with an orthogonality regularization loss $\mathcal{L}_{\text{orth}}$.\\  
Specifically, $\mathcal{L}_f$ is defined using cosine similarity between the estimated appearances $\hat{X}_t$ and the ground-truth embeddings $X_t$, forming an $n \times n$ similarity matrix $M$ that is scaled by a temperature parameter $\tau_f$. The diagonal entries correspond to matching object pairs, while off-diagonal entries correspond to mismatched object pairs. A cross-entropy objective is used to maximize similarity along the diagonal while minimizing similarity between different objects.
\begin{equation}
    \begin{aligned}
    M &= \frac{1}{\tau_f} (\hat{X}X^\top)\\
\mathcal{L}_{f} &= -\frac{1}{N} \sum_{i=1}^{N} \log \left( \frac{\exp(M_{ii})}{\sum_{j=1}^{N} \exp(M_{ij})} \right)
\end{aligned}
\end{equation}

To encourage Polycepta to learn well-structured and disentangled object-centric appearance states $H_t$ that support accurate appearance estimation, we employ an orthogonality regularization loss $\mathcal{L}_{\text{orth}}$, scaled by a temperature parameter $\tau_h$. This loss constructs a similarity matrix $M_{\text{orth}}$ from the appearance state, penalizes off-diagonal similarities while excluding the diagonal elements, thereby promoting distinct, non-redundant appearance-state representations across objects.
\begin{equation}
    \begin{aligned}
    M_{orth} &= \frac{1}{\tau_h} (HH^\top)\\
    \mathcal{L}_{orth} &= \frac{1}{N} \| M_{orth} \odot (\mathbf{1}\mathbf{1}^\top - I) \|_F
\end{aligned}
\end{equation}

Since the contrastive feature loss $\mathcal{L}_f$ operates in feature space while the orthogonality loss $\mathcal{L}_{\text{orth}}$ operates in state space, their magnitudes can differ substantially. To address this scale mismatch, we adopt a dynamic loss-balancing mechanism. The total training objective is defined as:$$\mathcal{L}_{total} = \frac{\mathcal{L}_{sim}}{\mu_{\text{sim}} + \epsilon} + \frac{\mathcal{L}_{orth}}{\mu_{\text{orth}} + \epsilon}$$where $\epsilon$ is a small constant for numerical stability. The normalization factors $\mu_{\text{sim}}$ and $\mu_{\text{orth}}$ are computed using an exponential moving average~\cite{6252962} of the corresponding loss magnitudes during training:$$\mu_{k, t} = \alpha \mu_{k, t-1} + (1 - \alpha) \bar{\mathcal{L}}_{k, t}$$where $\bar{\mathcal{L}}_{k,t}$ denotes the mean value of the $k$-th loss term within the current batch and $\alpha = 0.99$ is the momentum coefficient. This normalization stabilizes the training and removes the need for manual tuning of loss weights.
\section{Integration into MOT Benchmarks: Adaptive Gated Fusion}
\label{subsec:37_integration}

Polycepta is designed as a modular appearance-estimation component that can be integrated into existing tracking-by-detection pipelines, as illustrated in Figure~\ref{fig:integration}. Polycepta is incorporated in a manner analogous to the KF motion estimator: it receives matched object pairs from the association stage and recursively updates appearance states to estimate future appearance representations. This design allows Polycepta to enhance tracking and data association by providing temporally structured appearance predictions without modifying the underlying motion model or association strategy.

Concretely, the association cost matrix in the tracking benchmarks is augmented to fuse trajectory-based cues with Polycepta’s appearance-based similarity, while the original association mechanism remains unchanged.

Given the trajectory cost matrix $\mathbf{C}_k$ from a tracking benchmark, we construct an appearance-based association matrix from the cosine similarity between the current appearance features $\mathbf{X}_t \in \mathbb{R}^{n \times d_f}$ and the predicted appearance features from Polycepta $\hat{\mathbf{X}}_t \in \mathbb{R}^{m \times d_f}$:
\begin{equation}
\mathbf{M}_p = \mathbf{X}_t \hat{\mathbf{X}}_t^\top.
\end{equation}
An element-wise clamp is applied to enforce a minimum appearance-similarity threshold $\beta_p$ on the appearance similarity matrix:
\begin{equation}
\mathbf{M}_p' = \max(\mathbf{M}_p, \beta_p).
\end{equation}
Similarly, a trajectory threshold $\beta_k$ is applied to the trajectory cost matrix:
\begin{equation}
\mathbf{C}_k' = \min(\mathbf{C}_k, \beta_k).
\end{equation}
In tracking benchmarks that employ KF in 3D, lower values in $\mathbf{C}_k$ (Euclidean distance) indicate higher correspondence between predicted trajectories and detections, whereas higher values in $\mathbf{M}_p'$ (Cosine similarity) indicate stronger appearance similarity. To make the two costs compatible, we convert the similarity matrix $\mathbf{M}_p'$ into an appearance cost matrix $\mathbf{C}_p'$ by subtracting it from an all-ones matrix:
\begin{equation}
\mathbf{C}_p' = \mathbf{1}_n \mathbf{1}_m^\top - \mathbf{M}_p',
\end{equation}
where $n$ is the number of current detections and $m$ is the number of active tracks.
To ensure fusion compatibility, the appearance cost $\mathbf{C}_p'$ and trajectory cost $\mathbf{C}_k'$ are mapped to a common scale $[0, 1]$ by normalizing each matrix with respect to the bounded thresholds $\beta_p$ and $\beta_k$, respectively:
\begin{equation}
\tilde{\mathbf{C}}_k = \frac{\mathbf{C}_k'}{\beta_k}, 
\qquad 
\tilde{\mathbf{C}}_p = \frac{\mathbf{C}_p'}{1 - \beta_p}.
\end{equation}
Since the trajectory association matrix $\tilde{\mathbf{C}}_k$ and the appearance association matrix $\tilde{\mathbf{C}}_p$ originate from different sensing modalities or detection pipelines, the availability of detections can be asymmetric. This commonly arises from differences in the field of view (FoV) between LiDAR and camera sensors, or from using different detectors. To account for this, the final association cost $\mathbf{C}_{\text{fuse}}$ is obtained via a selective fusion mechanism that dynamically weights the trajectory and appearance modalities using weight matrices $\mathbf{W}_k$ and $\mathbf{W}_p$.\\
For each observation $i$ in a batch of $n$ objects, we define a binary availability vector $\boldsymbol{\gamma}_i = [\gamma_{k,i}, \gamma_{p,i}]$, where $\gamma \in \{0, 1\}$ indicates the existence of a valid trajectory estimate or appearance feature, respectively. The fusion weights for the $i$-th observation are determined by the following gating logic:
\begin{equation}
w_{k,i}, w_{p,i} =
\begin{cases}
\alpha, \; 1 - \alpha, & \text{if } \gamma_{k,i} \cdot \gamma_{p,i} = 1 \quad \text{(Bi-modal fusion)}, \\
1, \; 0, & \text{if } \gamma_{k,i} > \gamma_{p,i} \quad \text{(Trajectory gating)}, \\
0, \; 1, & \text{if } \gamma_{k,i} < \gamma_{p,i} \quad \text{(Appearance gating)}.
\end{cases}
\end{equation}
The global fused association matrix is then computed via element-wise  fusion, Hadamard product~\cite{10981717}:
\begin{equation}
\mathbf{C}_{\text{fuse}} = \mathbf{W}_k \odot \tilde{\mathbf{C}}_k + \mathbf{W}_p \odot \tilde{\mathbf{C}}_p,
\end{equation}
where $\alpha$ is a tunable prior that balances the influence of trajectory estimation and visual appearance when both modalities are available. This formulation allows Polycepta to provide appearance-state estimates as an auxiliary association cue while preserving the original tracking framework and assignment strategy.

\section{Results}
\vspace{-0.1cm}
\subsection{Experimental Setup}
\label{subsec:exp_setup}

Polycepta is implemented from scratch in PyTorch. All experiments are conducted on a mobile workstation equipped with an AMD Ryzen 9 processor, 32,GB of system memory, and a single NVIDIA RTX 3080 Laptop GPU with 16,GB of VRAM.

To evaluate the impact of appearance-state estimation in tracking-by-detection systems, Polycepta is integrated into our motion-based 3D tracking framework, RobMOT~\cite{robmot}, and the external motion-based 2D tracking benchmark FastTrack~\cite{fasttrack}. Since RobMOT is implemented in C++, Polycepta is compiled to C++17 via TorchScript. Experiments are conducted on the KITTI, WOD, and MOT17 benchmarks. Results are obtained using the official evaluation tools~\cite{eval_1,eval_2} or by submitting directly to the corresponding leaderboards. Detailed benchmark-specific integration settings are described in Section~\ref{subsec:42_kitti_comp}.

\subsection{Appearance Estimation Improves During Inference}
\begin{figure}[!t]
\centering
\includegraphics[width=\linewidth]{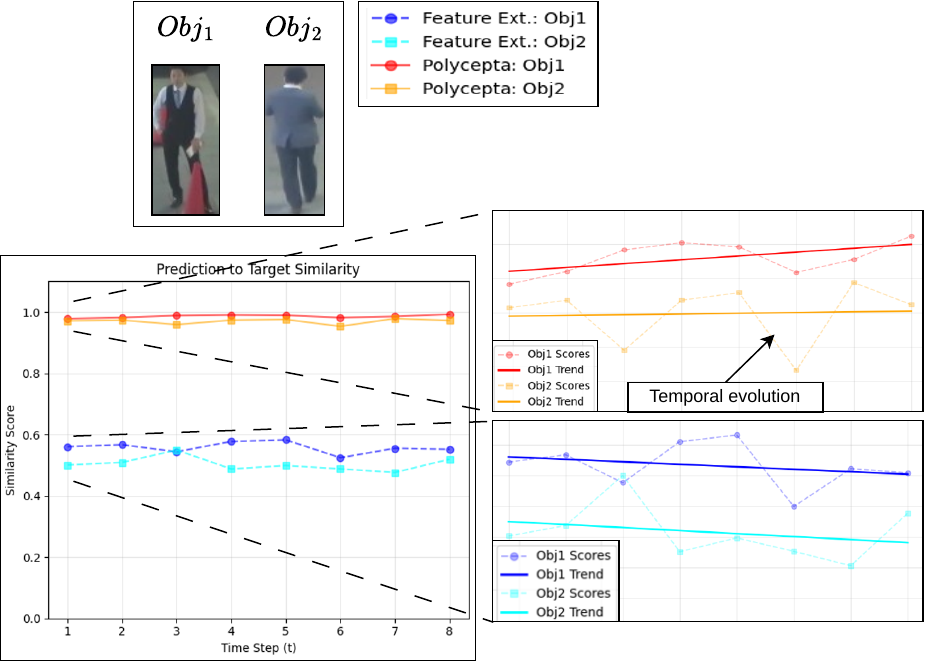}
\caption{A qualitative feature quality comparison between the traditional Re-ID and Polycepta for two pedestrians across consecutive frames. The similarity of the appearance embeddings to the next-frame appearance from MobileNetV3 (Re-ID) is highlighted in light and dark blue. Meanwhile, Polycepta's appearance embeddings are highlighted in red and orange.}
\label{fig:ablation_temp}
\vspace{-0.2cm}
\end{figure}

A central hypothesis of Polycepta is that appearance estimation quality should improve as appearance states accumulate information over time. To evaluate this hypothesis, we conduct both qualitative and quantitative studies comparing Polycepta against a conventional Re-ID. These experiments examine how appearance estimation evolves during inference and analyze the contribution of the proposed architecture components.

\subsubsection{Qualitative Appearance-State Evolution}
\label{subsec:temp_memory_eval}
The first study is conducted on the MOT17 dataset for two pedestrian objects, as shown in Figure~\ref{fig:ablation_temp}. The purpose of this study is to visualize how appearance estimation evolves during inference and to compare this behavior with a conventional Re-ID formulation.

Initially, the MobileNetV3 for feature embeddings is trained separately for five epochs using the contrastive learning loss function $L_f$ (Figure~\ref{fig:training_pipeline}), serving as a traditional Re-ID model. Similarly, Polycepta is trained on the same dataset for the same number of epochs. The experiment is conducted by selecting two pedestrians from the MOT17 streams and evaluating their appearance association across consecutive frames. 

First, the trained MobileNetV3 (Re-ID) is used to extract appearance embeddings, which are then associated with the next appearance embedding using a similarity score recorded in Figure~\ref{fig:ablation_temp}, marked in dark and light blue. Second, the trained Polycepta is used to replace the Re-ID method for appearance association evaluation, as shown in Figure~\ref{fig:ablation_temp}, marked in red and orange.

While the similarity scores produced by the Re-ID baseline gradually decrease over time, Polycepta exhibits the opposite behavior: appearance estimates become increasingly similar to future observations as appearance states accumulate information. This qualitative result provides initial evidence that recursive appearance-state estimation can improve appearance quality during inference.

\subsubsection{Quantitative Appearance-State Evolution}
\begin{figure}[!t]
\centering
\includegraphics[width=\linewidth]{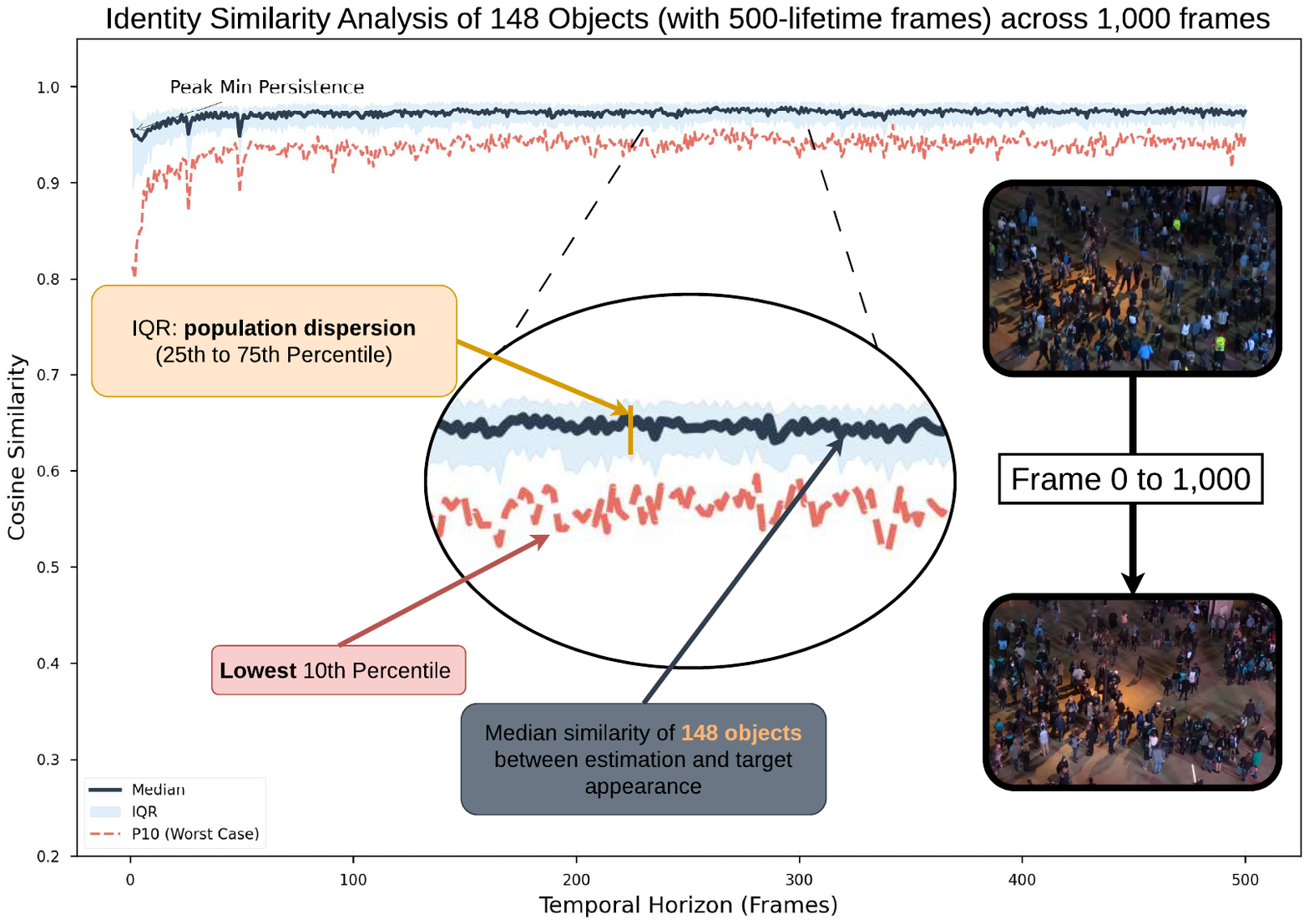}
\caption{A visualization of the conducted quantitative experiments for Polycepta on 148 objects with a lifetime of 500 frames across 1,000 frames in MOT20. The blue line shows the median (50th percentile) of the similarity scores between the next appearance estimates and the targets across objects. The red line shows the 10th percentile of the similarity scores, indicating challenging visual conditions and occlusions. The dispersion of the similarity scores is shown by the 20th to 75th percentile, highlighted in light blue.}
\label{fig:identity_exp}
\vspace{-0.2cm}
\end{figure}

To quantitatively evaluate appearance estimation during inference, we conduct a long-horizon experiment on MOT20~\cite{MOTChallenge20}, a highly crowded benchmark, involving 148 objects observed over 1,000 consecutive frames, as shown in Figure~\ref{fig:identity_exp}.

At each frame, the cosine similarity between the predicted appearance embedding of the object and its next-frame appearance is computed, and the median 50th percentile across objects is used as a representative of the method's appearance estimation quality, as shown by the dark blue line in Figure~\ref{fig:identity_exp}. 

The dispersion (IQR) of the similarity scores for the objects at each frame is shown as the 25th-75th percentile, highlighted in light blue in Figure~\ref{fig:identity_exp}. The challenging cases, the lowest 10th percentile similarity scores, are marked in red in Figure~\ref{fig:identity_exp}.

\begin{table*}[!t]
\centering
\caption{Quantitative evaluation of appearance estimation quality during inference and the contribution of Polycepta components, including the conventional Re-ID MobileNetV3 (MV3), for 148 objects with a lifetime of 500 frames across 1,000 frames in a crowded pedestrian bird-eye-view condition. \textbf{Bold} results indicate the highest performance across the metrics. The \textcolor{red}{red color $\downarrow$} indicates a reduction in appearance similarity over time, whereas the \textcolor{blue}{blue color $\uparrow$} indicates improvement.}
\label{tab:identity_table}

\begin{tabular}{|c|ccc|cc|cc|c|c|}
\hline
\multicolumn{4}{|c|}{\textbf{Architecture}} & \multicolumn{2}{c|}{\textbf{P50}} & \multicolumn{2}{c|}{\textbf{P10}} & \textbf{margin$\uparrow$}& \textbf{IDSw$\downarrow$} \\
MV3 & Polycepta & VRR & U-Gate & Start & End & Start & End && \\ \hline
\cmark &        &        &        & 0.493 & \textcolor{red}{0.472$\downarrow$} & 0.408 & \textcolor{red}{0.403$\downarrow$} & 0.045 & 9792\\
\cmark & \cmark &        &        & 0.950 & \textcolor{blue}{0.969$\uparrow$} & 0.832 & \textcolor{blue}{0.933$\uparrow$} & 0.290& 247 \\
\cmark & \cmark & \cmark &        & 0.951 & \textcolor{blue}{0.971$\uparrow$} & 0.798 & \textcolor{blue}{0.939$\uparrow$} & 0.288&222 \\
\cmark & \cmark &  &    \cmark    & 0.955 & \textcolor{blue}{0.975$\uparrow$} & 0.823 & \textcolor{blue}{0.947$\uparrow$} & 0.250&204 \\
\cmark & \cmark & \cmark & \cmark & 0.958 & \textcolor{blue}{0.975$\uparrow$} & 0.850 & \textcolor{blue}{0.949$\uparrow$} & 0.30 &\textbf{184}\\ 
\hline
\end{tabular}
\vspace{-0.3cm}
\end{table*}

The study involves the traditional feature extraction (Re-ID) as a standalone MobileNetV3, similar to the study in Section~\ref{subsec:temp_memory_eval}, and each module of Polycepta. Each component is trained for five epochs. 

The evaluation metrics used in this study are the start and end median similarity scores (50th percentile), which represent the median similarity scores at the beginning and end of the temporal horizon, as shown in Figure~\ref{fig:identity_exp} (the blue line). In addition, the start and end similarity scores for the challenging cases (the lowest 10th percentile) are used to measure the capability to handle scenarios with frequent occlusions, as shown in the red line in Figure~\ref{fig:identity_exp}.

To quantify the discrimination quality of the generated appearance embeddings and evaluate how distinct the target identity is from potential look-alikes, we introduce the \textit{discrimination margin} ($M$). Let $D_1$ denote the highest cosine similarity corresponding to the correct appearance estimation, and let $D_2$ denote the second-highest similarity score among all remaining candidate appearances. The margin is mathematically defined as:

\begin{equation}
M = D_1 - D_2
\end{equation}

A larger margin ($M \uparrow$) indicates superior discrimination capability, implying that the model confidently segregates the true identity from the closest false match, thereby mitigating the risk of identity switches (IDSw).

The first comparison, as shown in Table~\ref{tab:identity_table}, evaluates a conventional Re-ID, MobileNetV3~\cite{mobilenet}, against Polycepta without the inclusion of VRR and the selective appearance-state U-Gate (From Section~\ref{subsec:ocss}). 

\textbf{The 1st and 2nd row (ReID vs Polycepta):} The quantitative results support the phenomenon visualized in Figure~\ref{fig:ablation_temp}. MobileNetV3 (Re-ID) shows a downward trend in the 50th-percentile similarity score across the 148 objects (0.493→ 0.472). This behavior is also observed in the worst case (the lowest 10th percentile), as shown in Table~\ref{tab:identity_table}. On the other hand, Polycepta shows a strong start, with a 50th-percentile similarity score of 0.950, which then increases over time to 0.969. Unlike the Re-ID baseline, which exhibits a gradual reduction in appearance quality over time, Polycepta demonstrates a consistent improvement throughout the temporal horizon. Moreover, the discrimination capability between objects (margin) was enhanced by Polycepta (0.045→0.290) with a noticeable reduction in identity switch between objects (9792→247).  

 \textbf{The 3rd row (VRR):} The inclusion of the VRR module (Section~\ref{subsec:vssp}), which correlates current and historical appearances for feature generation, further reduces IDSw ($247 \rightarrow 222$) and improves final appearance quality ($\text{P50}: 0.969 \rightarrow 0.971 \mid \text{P10}: 0.933 \rightarrow 0.939$). However, VRR slightly degrades the initial appearance quality at the track's start, as limited appearance information has accumulated during the early stages, which VRR relies on.

\textbf{The 4th row (U-Gate):} In contrast, the integration of the U-Gate (Section~\ref{subsec:ocss}) maintains robust appearance quality at earlier times by dynamically prioritizing the incoming appearance over an sparsely initialized appearance state. The U-Gate yields a notable improvement in final appearance quality ($\text{P50}: 0.969 \rightarrow 0.975 \mid \text{P10}: 0.933 \rightarrow 0.947$) alongside a significant reduction in IDSw ($247 \rightarrow 204$). However, incorporating the U-Gate alone into Polycepta reduces the system's overall discrimination capability, compressing the margin from $0.290$ to $0.250$. This narrower margin suggests a higher vulnerability to mismatches under ambiguous or closely matching conditions.

\textbf{The 5th row (Full Configuration):} By simultaneously incorporating both VRR and the U-Gate into Polycepta, the architecture overcomes the individual limitations of each isolated module. This full configuration shows consistent improvements across all metrics, achieving the lowest IDSw ($184$) and recovering a high discrimination margin ($0.300$), effectively demonstrating the complementary and incremental contributions of both components.

Collectively, the qualitative and quantitative experiments demonstrate that Polycepta differs fundamentally from frame-wise appearance descriptors. While the quality of Re-ID features degrades over time, Polycepta progressively improves appearance estimation quality as appearance states evolve (Figure~\ref{fig:ablation_temp}). 
\begin{table*}[!t]
\centering
\caption{ Evaluation of online MOT methods on the KITTI test dataset. Bold \textcolor{red}{\textbf{ red}} indicates the highest value in each metric, and \textcolor{blue}{\textbf{blue}}  indicates the second-highest. (*) denotes methods using CasA 3D detector~\cite{casA_detector}. Underlined values highlight the best scores.}
\label{table:kitti_benchmarks}
\begin{tabular}{|llccccccc|}
\hline
\textbf{Method} & \textbf{Venue} & \textbf{Input} & \textbf{HOTA} & \textbf{DetA} & \textbf{AssA} & \textbf{MOTA} & \textbf{IDSw} & \textbf{FN} \\ \hline
\rowcolor[gray]{0.95} \multicolumn{9}{|l|}{\textit{3D-based Paradigms}} \\ \hline
PC3TMOT~\cite{pc3t} & T-ITS '21 & 3D & 77.80 & 74.57 & 81.59 & 88.81 & 225 & 2810 \\
PolarMOT~\cite{polarmot} & ECCV '22 & 3D & 75.16 & 73.94 & 76.95 & 85.08 & 462 & 2668 \\
EAFFMOT~\cite{eaffmot} & Sig. Proc. '24 & 3D & 72.28 & 71.97 & 73.08 & 84.77 & 107 & 3946 \\
UG3DMOT~\cite{ug3dmot} & Sig. Proc. '24 & 3D & 78.60 & 76.10 & 82.28 & 87.98 & 40 & 2993 \\
MCTrack~\cite{mctrack}*&IROS '25&3D&78.12&73.24&84.01&83.89&63&3364\\
RobMOT*~\cite{robmot,nagy} & T-ITS '25 & 3D & 80.83 & 76.99 & 85.50 & 89.83  & \textcolor{blue}{\textbf{6}} & 1544 \\  \hline
\rowcolor[HTML]{EFEFEF} \textbf{RobMOT*~\cite{robmot,nagy} (w/ Polycepta)} &  & \textbf{3D} & \textcolor{blue}{\textbf{81.28}} &  \textcolor{blue}{\textbf{77.75}} &  \textcolor{red}{\textbf{85.61}} &  \textcolor{blue}{\textbf{90.84}} &  \textcolor{red}{\textbf{5}} &  \textcolor{blue}{\textbf{1543}} \\ \hline
\rowcolor[gray]{0.95} \multicolumn{9}{|l|}{\textit{Fusion-based Paradigms}} \\ \hline
JMODT~\cite{jmodt} & IROS '21 & 2D+3D & 70.73 & 73.45 & 68.76 & 85.35  & 350 & 3438 \\
DeepFusionMOT~\cite{deepfusion} & RA-L '22 & 2D+3D & 75.46 & 71.54 & 80.06 & 84.64& 84 & 4601 \\
StrongFusion~\cite{strongfusionmot} & Sensors '22 & 2D+3D & 75.65 & 72.08 & 79.84 & 85.53 & 58 & 4658 \\
BcMODT~\cite{bcmodt} & Rem. Sens. '23 & 2D+3D & 71.00 & 73.62 & 69.14 & 85.48 & 381 & 3353 \\
PNAS-MOT~\cite{peng2024pnasmot} & RA-L '24 & 2D+3D & 67.32 & 77.69 & 58.99 & 89.59 & 751 & 2261 \\
MMF-JDT*~\cite{10777493}& RA-L '25 & 2D+3D & 79.52 & 75.83 & 84.01 & 88.06  & 37 & 1751 \\
Hui Li et al.~\cite{10771607}&T-IV '25&2D+3D&76.12&-&-&88.43&-&1622\\ 
\rowcolor[HTML]{EFEFEF} \textbf{RobMOT*~\cite{robmot,nagy} (w/ Polycepta)} & \textbf{-} & \textbf{2D + 3D} & \textcolor{red}{\textbf{81.35}} & \textcolor{red}{\textbf{78.95}} & \textcolor{blue}{\textbf{84.41}} & \textcolor{red}{\textbf{91.80}}  & 76 & \textcolor{red}{\textbf{1127}} \\
\hline
\hline
\rowcolor[HTML]{F2F9FF}
\textbf{RobMOT~\cite{robmot,nagy} (w/ Polycepta)} & \textbf{-} & \textbf{2D + 3D} & \underline{82.05} & \underline{80.16} & \underline{84.56} & \underline{92.27}  & 73 & \underline{823}\\
\hline
\end{tabular}%
\vspace{-0.2cm}
\end{table*}

\vspace{-0.1cm}
\subsection{Performance in Motion-Based MOT Frameworks}
\label{subsec:42_kitti_comp}
We evaluate Polycepta under two integration settings. In the first setting, Polycepta is integrated using a single detector (2D or 3D). For 3D detectors, detections are projected into the image plane prior to appearance estimation. In the second setting, Polycepta is evaluated using a 2D+3D detector configuration, where YOLOv8 is fused with the existing 3D detector.

In this experiment, Polycepta is integrated into the recent benchmark RobMOT~\cite{robmot,nagy}, which employs a 3D CasA~\cite{casA_detector} detector on the KITTI~\cite{geiger2012kitti} dataset, following the \textbf{first setting (LiDAR to camera projection)} described in Section~\ref{subsec:exp_setup}.

The integration consistently improves tracking accuracy and association quality across all reported metrics. In particular, MOTA increases from 89.83\% to 90.84\%, while the number of identity switches decreases from 6 to 5, representing the lowest IDSw among the reported methods.

In the \textbf{second setting} from Section~\ref{subsec:exp_setup}, the YOLOv8 2D detector~\cite{yolov8_ultralytics} is fused with the 3D CasA~\cite{casA_detector} detector in RobMOT~\cite{robmot,nagy}. As shown in Table~\ref{table:kitti_benchmarks}, this fused setup outperforms the reported fusion-based baselines, achieving approximately a 4\% improvement in MOTA.

When integrated with RobMOT using the VirConv~\cite{VirConvDet_CVPR_23} detector, Polycepta achieves state-of-the-art performance on the KITTI benchmark, reaching a MOTA of 92.27\%, as summarized in Table~\ref{table:kitti_benchmarks}.

These experiments demonstrate that Polycepta's appearance-state estimates yield consistent tracking improvements when integrated into motion-tracking frameworks. The gains are observed under both single-detector and multi-detector configurations, indicating that the proposed appearance-state formulation complements existing motion-estimation pipelines without requiring modifications to the underlying tracking framework. Furthermore, the consistent improvements across integration settings suggest that the benefits of appearance-state estimation extend beyond a specific association configuration.
\vspace{-0.1cm}
\subsection{Performance Across 2D and 3D Tracking Datasets}
\begin{table*}[!b]
\vspace{-0.3cm}
\centering
\caption{ Performance evaluation across MOT17, KITTI, and WOD using motion-centric MOT frameworks with unified detector configurations.}
\label{tab:all_datasets}
\resizebox{\textwidth}{!}{%
\begin{tabular}{|llc|ll|lll|l|}
\hline
\rowcolor[gray]{0.8}
\textbf{Method} & \textbf{Venue} & \textbf{Input} &
\multicolumn{2}{c|}{\textbf{Tracking Accuracy}} &
\multicolumn{3}{c|}{\textbf{Association Quality}} &
\textbf{IDSw$\downarrow$} \\ \cline{4-9}

\rowcolor[gray]{0.8}
 &  &  &
\textbf{HOTA$\uparrow$} &
\textbf{MOTA$\uparrow$} &
\textbf{AssA$\uparrow$} &
\textbf{AssP$\uparrow$} &
\textbf{AssR$\uparrow$} &
 \\ \hline

\rowcolor[gray]{0.95} \multicolumn{9}{|l|}{\textbf{MOT17 Benchmark (2D) – Pedestrians}} \\ \hline
TrackTrack~\cite{tracktrack} & CVPR '25 & 2D
& 66.90 & 79.36  & 67.83 & 82.42 & 73.19 & \textbf{254} \\

FastTracker~\cite{fasttrack} & ArXiv '25 & 2D
& 73.04 & 88.34  & 69.97 & 79.93 & 77.30 & 567 \\
 \hline
\rowcolor[HTML]{F2F9FF}
\textbf{FastTracker~\cite{fasttrack} (w/ Polycepta)} & & \textbf{2D}
& \textbf{75.12} \textcolor{blue}{(+2.08$\uparrow$)} & \textbf{88.69} \textcolor{blue}{(+0.35$\uparrow$)} & \textbf{73.31}\textcolor{blue}{(+3.34$\uparrow$)} & \textbf{84.05} \textcolor{blue}{(+4.12$\uparrow$)} & \textbf{79.52} \textcolor{blue}{(+2.22$\uparrow$)}  & 382 \textcolor{blue}{(-185$\downarrow$)} \\ \hline

\rowcolor[gray]{0.95} \multicolumn{9}{|l|}{\textbf{KITTI Benchmark (2D) – Cars}} \\ \hline
MCTrack~\cite{mctrack} & IROS '25 & 3D
& 80.09 & 81.24  & 85.70 & \textbf{90.81} & 90.93 & 7 \\

RobMOT~\cite{robmot,nagy} & T-ITS '25 & 3D
& 83.06 & 89.57 & 85.77 & 90.04 & 91.06 & \textbf{1} \\
 \hline
\rowcolor[HTML]{F2F9FF}
\textbf{RobMOT (w/ Polycepta)} && \textbf{3D + 2D}
& \textbf{84.54} \textcolor{blue}{(+1.48$\uparrow$)} & \textbf{92.18} \textcolor{blue}{(+2.61$\uparrow$)} & \textbf{86.80} \textcolor{blue}{(+1.03$\uparrow$)} & 90.53 \textcolor{blue}{(+0.49$\uparrow$)} & \textbf{91.49} \textcolor{blue}{(+0.43$\uparrow$)} & 3 \textcolor{red}{(+2$\uparrow$)} \\ \hline

\rowcolor[gray]{0.95} \multicolumn{9}{|l|}{\textbf{Waymo Open Dataset Benchmark (3D) – Vehicles}} \\ \hline
MCTrack~\cite{mctrack} & IROS '25 & 3D
& 55.48 & 42.61  & 66.88 & 73.37 & \textbf{75.68} & 811 \\

RobMOT~\cite{robmot,nagy} & T-ITS '25 & 3D
& \textbf{57.33} & 52.22 & 67.81 & 75.61 & 74.83 & 501 \\
 \hline
\rowcolor[HTML]{F2F9FF}
\textbf{RobMOT~\cite{robmot,nagy} (w/ Polycepta)} & & 3D + 2D
& 57.31 \textcolor{red}{(-0.02$\downarrow$)} & \textbf{52.46} \textcolor{blue}{(+0.24$\uparrow$)} & \textbf{68.05}\textcolor{blue}{(+0.24$\uparrow$)} & \textbf{75.72} \textcolor{blue}{(+0.11$\uparrow$)} & 74.05 \textcolor{red}{(-0.78$\downarrow$)} & \textbf{444} \textcolor{blue}{(-57$\downarrow$)} \\ \hline
\end{tabular}%
}
\vspace{-0.2cm}
\end{table*}
Having established that appearance estimation quality improves during inference, we next evaluate whether the resulting appearance-state formulation generalizes across diverse tracking scenarios, datasets, and sensing modalities. 

First, Polycepta is integrated into the FastTracker~\cite{fasttrack} benchmark (currently ranked 1st on the MOT17~\cite{mot17} leaderboard) using the \textbf{first setting} described in Section~\ref{subsec:42_kitti_comp}. No additional detector is required, as FastTracker already employs a 2D YOLOvx~\cite{yolox2021} detector. Table~\ref{tab:all_datasets} summarizes the highest-performing benchmarks for each dataset, including RobMOT~\cite{robmot,nagy}, MCTrack~\cite{mctrack}, FastTracker~\cite{fasttrack}, and TrackTrack~\cite{tracktrack}.

\textbf{MOT17 Benchmark (2D) Table~\ref{tab:all_datasets}: } Integrating Polycepta into FastTracker~\cite{fasttrack} reduces identity switches from 567 to 382 while improving all association metrics, including AssA (+3.34\%), AssP (+4.12\%), and AssR (+2.22\%). These gains translate into a 2.08\% improvement in HOTA, indicating stronger association quality and more stable identity preservation.

An additional experiment is conducted using the top-performing benchmarks, RobMOT~\cite{robmot,nagy} and MCTrack~\cite{mctrack}, on the KITTI~\cite{geiger2012kitti} and WOD~\cite{sun2020waymo} validation datasets. The Polycepta YOLOv8~\cite{yolov8_ultralytics} fusion settings are applied, as explained in Section~\ref{subsec:42_kitti_comp} (\textbf{second setting}), while the 3D CasA detector~\cite{casA_detector} is unified across all benchmarks in Table~\ref{tab:all_datasets}. 

\textbf{KITTI Benchmark (2D) Table~\ref{tab:all_datasets}}: 
The integration of Polycepta to RobMOT shows systematic improvements in the tracking metric metrics HOTA and MOTA by 1.48\% and 2.61\%, respectively. In addition to improvement in all association quality metrics, the same behavior was observed in MOT17 with FastTracker. The increase in IDSw (by 2 objects) is due to the fusion of an additional detector (YOLOv8), leading to detections unannotated in the KITTI ground truth and false positives. 

\textbf{WOD Benchmark (3D Point Cloud) Table~\ref{tab:all_datasets}}: 
On WOD, Polycepta primarily improves association quality, reducing identity switches from 501 to 444 despite operating on image observations, while evaluation is performed in the 3D point-cloud domain. This result suggests that the appearance-state estimates remain beneficial even when the final tracking metrics are evaluated on a 3D LiDAR point cloud.

Across MOT17, KITTI, and WOD, Polycepta consistently improves association quality and reduces identity switches under both 2D and 3D tracking settings. 

\subsection{Detector Generalization}
\begin{table*}[!t]
\centering
\caption{ Comparison of RobMOT~\cite{robmot,nagy} performance with and without Polycepta across five detectors. Best results for each detector are highlighted in \textbf{bold}.}
\label{tab:diff_detectors}

\begin{tabular}{@{}|l|l|l|l|l|l|@{}}
\hline
Detector & Config & HOTA $\uparrow$ & MOTA $\uparrow$ & MT $\uparrow$ & ML $\downarrow$ \\ \hline

VirConv~\cite{VirConvDet_CVPR_23} & RobMOT~\cite{robmot,nagy} & 86.57 & 91.60 & 159 & 14 \\
 & w/ Polycepta & \textbf{86.74} \textcolor{blue}{(+0.17$\uparrow$)} & \textbf{93.05} \textcolor{blue}{(+1.45$\uparrow$)} & \textbf{174} \textcolor{blue}{(+15$\uparrow$)} & \textbf{3} \textcolor{blue}{(-11$\downarrow$)} \\ \hline

PV-RCNN~\cite{pv-rcnn_detector} & RobMOT~\cite{robmot,nagy} & 80.22 & 87.14 & 155 & 11 \\
 & w/ Polycepta & \textbf{80.96} \textcolor{blue}{(+0.74$\uparrow$)} & \textbf{90.92} \textcolor{blue}{(+3.78$\uparrow$)} & \textbf{170} \textcolor{blue}{(+15$\uparrow$)} & \textbf{6} \textcolor{blue}{(-5$\downarrow$)} \\ \hline

CasA~\cite{casA_detector} & RobMOT~\cite{robmot,nagy} & 83.06 & 89.57 & 166 & 11 \\
 & w/ Polycepta & \textbf{84.54} \textcolor{blue}{(+1.48$\uparrow$)} & \textbf{92.18} \textcolor{blue}{(+2.61$\uparrow$)} & \textbf{174} \textcolor{blue}{(+8$\uparrow$)} & \textbf{8} \textcolor{blue}{(-3$\downarrow$)} \\ \hline

PointRCNN~\cite{point_rcnn_detector} & RobMOT~\cite{robmot,nagy} & 78.22 & 86.72 & 153 & 10 \\
 & w/ Polycepta & \textbf{79.79} \textcolor{blue}{(+1.57$\uparrow$)} & \textbf{90.70} \textcolor{blue}{(+3.98$\uparrow$)} & \textbf{168} \textcolor{blue}{(+15$\uparrow$)} & \textbf{7} \textcolor{blue}{(-3$\downarrow$)} \\ \hline

Second~\cite{second_detector} & RobMOT~\cite{robmot,nagy} & 78.91 & 85.85 & 155 & 12 \\
 & w/ Polycepta & \textbf{80.40} \textcolor{blue}{(+1.49$\uparrow$)} & \textbf{90.75} \textcolor{blue}{(+4.90$\uparrow$)} & \textbf{162} \textcolor{blue}{(+7$\uparrow$)} & \textbf{12} \textcolor{blue}{(+0)} \\ \hline
\end{tabular}
\vspace{-0.2cm}
\end{table*}
\begin{table*}[!b]
\vspace{-0.2cm}
\centering
\caption{ Generalization of Polycepta trained on a certain object-class and evaluated on an unseen different class.}
\label{tab:zero-shot}
\begin{tabular}{llc c c c c}
\toprule
\textbf{Train Class} & \textbf{Test Class} & \textbf{Method} 
& \textbf{MOTA}$\uparrow$ 
& \textbf{MOTP}$\uparrow$ 
& \textbf{MT}$\uparrow$ 
& \textbf{ML}$\downarrow$ \\ 
\midrule

Pedestrian (MOT17/20) & Pedestrian (MOT17/20) & FastTracker~\cite{fasttrack} 
& 93.05 & 91.46 & 174 & 3 \\
Vehicle (KITTI/WOD)  & Pedestrian (MOT17/20) & 
& 92.33 & 91.68 & 173 & 4 \\
\cmidrule(lr){4-7}
\textit{Performance Gap} & & 
& -0.72 & +0.22 & -1 & +1 \\

\midrule
Vehicle (KITTI/WOD)  & Vehicle (KITTI/WOD) & RobMOT~\cite{robmot,nagy} 
& 88.69 & 87.75 & 399 & 25 \\
Pedestrian (MOT17/20) & Vehicle (KITTI/WOD) & 
& 88.63 & 87.72 & 401 & 26 \\
\cmidrule(lr){4-7}
\textit{Performance Gap} & & 
& -0.06 & -0.03 & +2 & +1 \\

\bottomrule
\end{tabular}
\end{table*}

The previous experiments used the 3D CasA~\cite{casA_detector} detector on KITTI~\cite{geiger2012kitti} to maintain a unified detector across benchmarks. To assess robustness to detector choice and quantify performance variance after integrating Polycepta, we conduct additional experiments with multiple 3D detectors. Table~\ref{tab:diff_detectors} reports results using VirConv~\cite{VirConvDet_CVPR_23}, PV-RCNN~\cite{pv-rcnn_detector}, CasA~\cite{casA_detector}, PointRCNN~\cite{point_rcnn_detector}, and Second~\cite{second_detector} on the KITTI validation set. Since the objective is to evaluate tracking capability rather than detection quality, we additionally report Mostly Tracked (MT) and Mostly Lost (ML), which measure the number of objects that are consistently tracked throughout their lifespan or predominantly lost by the tracker, respectively.

The experiment demonstrates that integrating Polycepta consistently improves tracking performance across detectors of varying quality, including weaker 3D detectors such as PointRCNN~\cite{point_rcnn_detector} and SECOND~\cite{second_detector}. As shown in Table~\ref{tab:diff_detectors}, Polycepta substantially improves MOTA under low-quality detections, surpassing 90\% when the baseline margin exceeds 5\%, e.g., with the SECOND detector~\cite{second_detector}, where MOTA increases from 85.85\% to 90.75\%. The largest improvements are observed in these weaker detectors, suggesting that appearance-state estimation can compensate for noisier detection inputs and improve the robustness of association.

Even when paired with a strong detector such as VirConv~\cite{VirConvDet_CVPR_23}, Polycepta improves long-term tracking continuity. Specifically, ML decreases from 14 to 3, while MT increases from 159 to 174.

Overall, these results indicate that the appearance-state estimates produced by Polycepta remain effective across detector qualities, improving tracking continuity and reducing identity fragmentation even when detections are noisy or intermittent.

\subsection{Cross-Category Generalization Evaluation}
\label{subsec:zero-shot}

To evaluate Polycepta's ability to generalize beyond object categories observed during training, we conduct a cross-category evaluation in which training and testing are performed on disjoint object classes. In this experiment, Polycepta is trained on the vehicle class in the KITTI and WOD datasets, then evaluated on the pedestrian class in the MOT17/20 dataset, with the roles reversed. 

\textbf{Train on vehicle class (KITTI/Waymo), test on pedestrian class (MOT17/20); top section in Table~\ref{tab:zero-shot}}: The performance degradation relative to training on the target class is minimal, with a MOTA reduction of only 0.72\% and negligible changes in MT and ML.

\textbf{Train on pedestrian class (MOT17/20), test on vehicle class (KITTI/Waymo); bottom section in Table~\ref{tab:zero-shot}:} The performance gap remains negligible, with only a -0.06\% change in MOTA and a -0.03\% change in MOTP. The consistently small performance gap in both transfer directions indicates that the learned appearance-state representations are not strongly tied to any specific object category.

These results indicate that Polycepta learns appearance-state construction mechanisms that transfer across object categories. The small performance gap on unseen classes suggests that the model relies on general appearance-state evolution dynamics rather than memorizing category-specific appearance patterns observed during training. 

This behavior is consistent with the motivation behind the proposed state-erasure learning strategy, which encourages the model to learn how appearance states are constructed rather than memorizing object-specific training instances.

\section{Ablation Study}
\subsection{Orthogonality-Guided Appearance-State Learning}
\label{subsec:ablation_learning_loss}

\begin{table}[!t]
\centering
\caption{ Effect of the orthogonality loss $\mathcal{L}_{\text{orth}}$ on appearance-state separation and appearance estimation over 10 frames}
\label{tab:orthogonality_ablation}
\resizebox{\linewidth}{!}{

\begin{tabular}{@{}|llcc|@{}}
\hline
\textbf{Feature Stage} & \textbf{Metric} & \textbf{w/o $\mathcal{L}{orth}$} & \textbf{w/  $\mathcal{L}{orth}$} \\ \hline
Input Features         & Mean $\pm$ std  & $0.4039 \pm 0.0478$          & $0.5337 \pm 0.0452$         \\
                       & Range           & $[0.3256, 0.4823]$           & $[0.4412, 0.5960]$          \\
                       & Similarity at $10^{th}$           & $0.4288$                     & $0.5636$                    \\ \hline
Appearance State & Mean $\pm$ std  & $\mathbf{1.0000 \pm 0.0000}$ & $\mathbf{-0.5119 \pm 0.6351}$ \\
                       & Range           & $[1.0000, 1.0000]$           & $[-0.9962, 0.9720]$         \\
                       & Similarity at $10^{th}$        & $\mathbf{1.0000}$            & $\mathbf{-0.8051}$          \\ \hline
Estimated Appearance        & Mean $\pm$ std  & $0.2821 \pm 0.0671$          & $0.3154 \pm 0.0899$         \\
                       & Range           & $[0.1570, 0.3942]$           & $[0.1465, 0.4751]$          \\
                       & Similarity at $10^{th}$               & $0.3306$                     & $0.3280$                    \\ \hline
\end{tabular}
}
\vspace{-0.3cm}

\end{table}

We conduct an ablation study by removing the orthogonality loss $\mathcal{L}_{\text{orth}}$ and training the Polycepta for 5 epochs using only the contrastive loss $\mathcal{L}_f$. We then trace the evolution of two distinct pedestrians over 10 consecutive frames.

The analysis metrics:
\begin{itemize}
    \item \textbf{Input Features}: The cosine similarity between the appearance embeddings of the two pedestrians extracted by MobileNetV3-small.
    \item \textbf{Appearance State}: The cosine similarity between the appearance states of the two pedestrians.
    \item \textbf{Estimated appearance}: The cosine similarity between the estimated appearances of the two pedestrians.
\end{itemize}

Two key observations can be drawn from Table~\ref{tab:orthogonality_ablation}. 

First, the estimated appearance similarity between pedestrians is consistently lower than the similarity of the original input feature embedding used by Polycepta. The input feature similarity ranges from $[0.3256, 0.4823]$ without $\mathcal{L}_{\text{orth}}$ and $[0.4412, 0.5960]$ with $\mathcal{L}_{\text{orth}}$, while the estimated appearance similarity ranges from $[0.1570, 0.3942]$ and $[0.1465, 0.4751]$, respectively. This confirms that Polycepta enhances feature discriminability beyond what ReID provides. Lower appearance similarity directly implies reduced ambiguity during data association, thereby lowering the likelihood of identity mismatches.

When the orthogonality loss $\mathcal{L}_{\text{orth}}$ is removed, the appearance states of different objects collapse to nearly identical representations, as evidenced by a constant similarity of 1.0 across all measurements. This indicates that the appearance-state update process no longer preserves object-specific information, causing the model to rely primarily on the feature extractor rather than on the learned appearance states.

In contrast, employing $\mathcal{L}_{\text{orth}}$ encourages the formation of distinct appearance states. While the input feature similarity increases, the appearance-state similarity spans a wide range of $[-0.9962, 0.9720]$, with a strongly negative similarity at the last frame ($-0.8051$). This indicates that the appearance states of different objects are actively pushed apart over time. The strongly negative similarity observed at later frames indicates that appearance states remain highly discriminative even when the underlying input features become increasingly similar. Remarkably, enforcing $\mathcal{L}_{\text{orth}}$ results in comparable or lower appearance estimation similarity at the final frame ($0.3280$ vs.\ $0.3306$), despite higher input feature similarity. This demonstrates that orthogonality regularization prevents appearance-state collapse and enables Polycepta to construct robust and discriminative appearance states that improve identity preservation during association.

\subsection{State-Erasure Learning}
\begin{table}[!t]
\centering
\caption{Memory erasure evaluation on seen and unseen object classes after five training epochs.}
\label{tab:memory_erasure}
\begin{tabular}{l|cc|cc}
\toprule
 & \multicolumn{2}{c|}{\textbf{Seen Classes}} & \multicolumn{2}{c}{\textbf{Unseen Classes}} \\
\textbf{Method} & \textbf{margin$\uparrow$} & \textbf{IDSw$\downarrow$} & \textbf{margin$\uparrow$} & \textbf{IDSw$\downarrow$} \\
\midrule
w/o Erasure & \textbf{0.337} & 224 & 0.169 & 144 \\
w/ Erasure  & 0.300 & \textbf{184} & \textbf{0.188} & \textbf{118} \\
\bottomrule
\end{tabular}
\end{table}

To evaluate the contribution of the proposed state-erasure learning strategy, we train Polycepta with and without state erasure for five epochs and evaluate both models on seen and unseen object classes.  Table~\ref{tab:memory_erasure} summarizes the results of this experiment.

The model trained with state erasure consistently achieves lower ID-switch counts on both seen and unseen classes, with the largest improvement observed on unseen categories (144 → 118). Although the model trained without erasure attains a higher discrimination margin on seen classes (0.337), this advantage deteriorates on unseen classes, where the margin drops to 0.169. In contrast, the state-erasure model maintains a more stable margin while achieving the lowest ID-switch counts across both settings.

This behavior suggests that, without state erasure, the model increasingly relies on object-specific appearance information accumulated during training. In contrast, state erasure encourages the network parameters to learn appearance-state construction mechanisms that transfer more effectively to unseen categories. These results support the hypothesis that state erasure discourages memorization of training objects.



\begin{table}[!t]
\centering
\caption{ Computational throughput of the complete tracking pipeline.}
\label{tab:timing}
\begin{tabular}{@{}lc@{}}
\toprule
\textbf{Component} & \textbf{Throughput (Hz)} \\ \midrule
Feature Extraction (\textit{MobileNetV3-Small}) & 124.78 \\
Polycepta (Inference \& State Evolution)         & 377.16 \\ 
Object Association                               & 2713.25 \\ \midrule
\textbf{Full System Total}                       & \textbf{90.57} \\ 
\bottomrule
\end{tabular}
\vspace{-0.3cm}
\end{table}
\subsection{Efficiency Analysis}  
We evaluate Polycepta's computational throughput on the mobile workstation described in Section~\ref{subsec:exp_setup} by integrating it into the RobMOT~\cite{robmot,nagy} tracking framework. As reported in Table~\ref{tab:timing}, the proposed adaptive gated fusion association mechanism achieves a throughput of 2713.25 Hz, while Polycepta's appearance-state estimation module operates at 377.16 Hz. When combined with feature extraction and association, the complete tracking pipeline maintains a throughput of 90.57 Hz. These results demonstrate that Polycepta can provide online appearance-state estimation while satisfying real-time operating requirements for autonomous systems.
\section{Conclusion}

We presented Polycepta, an object-centric appearance-state estimation framework for multi-object tracking that reformulates appearance modeling as a recursive estimation problem rather than a frame-wise matching task. By constructing and continuously updating an appearance state for each tracked object, Polycepta enables future appearance representations to be estimated from accumulated observations, allowing appearance estimation quality to improve during inference as appearance states evolve. Extensive experiments demonstrated that this formulation consistently improves association quality and reduces identity switches when integrated into motion-based tracking-by-detection frameworks across KITTI, MOT17, and the Waymo Open Dataset, while maintaining real-time operation at 90.57 Hz. Furthermore, the proposed state-erasure learning strategy encourages the learning of category-agnostic appearance-state construction, enabling generalization to previously unseen object categories with only marginal performance degradation. Collectively, these results suggest that recursive appearance-state estimation provides an effective complement to motion-state estimation in tracking-by-detection systems, offering a practical and lightweight approach for improving long-term identity preservation in real-time multi-object tracking.

\bibliographystyle{IEEEtran}
\bibliography{egbib}

@ARTICLE{10777493,
  author={Wang, Xiyang and Fu, Chunyun and He, Jiawei and Huang, Mingguang and Meng, Ting and Zhang, Siyu and Zhou, Hangning and Xu, Ziyao and Zhang, Chi},
  journal={IEEE Robotics and Automation Letters}, 
  title={A Multi-Modal Fusion-Based 3D Multi-Object Tracking Framework With Joint Detection}, 
  year={2025},
  volume={10},
  number={1},
  pages={532-539},
  doi={10.1109/LRA.2024.3511438}}

@inproceedings{Samba,
 author = {Segu, Mattia and Piccinelli, Luigi and Li, Siyuan and Yang, Yung-Hsu and Van Gool, Luc and Schiele, Bernt},
 booktitle = {International Conference on Learning Representations},
 editor = {Y. Yue and A. Garg and N. Peng and F. Sha and R. Yu},
 pages = {30057--30070},
 title = {Samba: Synchronized Set-of-Sequences Modeling for Multiple Object Tracking},
 volume = {2025},
 year = {2025}
}

@INPROCEEDINGS{MeMOT,
  author={Cai, Jiarui and Xu, Mingze and Li, Wei and Xiong, Yuanjun and Xia, Wei and Tu, Zhuowen and Soatto, Stefano},
  booktitle={2022 IEEE/CVF Conference on Computer Vision and Pattern Recognition (CVPR)}, 
  title={MeMOT: Multi-Object Tracking with Memory}, 
  year={2022},
  volume={},
  number={},
  doi={10.1109/CVPR52688.2022.00792}}

@INPROCEEDINGS{MeMOTR,
  author={Gao, Ruopeng and Wang, Limin},
  booktitle={2023 IEEE/CVF International Conference on Computer Vision (ICCV)}, 
  title={MeMOTR: Long-Term Memory-Augmented Transformer for Multi-Object Tracking}, 
  year={2023},
  volume={},
  number={},
  pages={9867-9876},
  doi={10.1109/ICCV51070.2023.00908}}

@article{strongsort,
  author    = {Du, Yunhao and Zhao, Zhicheng and Song, Yang and Zhao, Yanyun and Su, Fei and Gong, Tao and Meng, Hongying},
  journal   = {IEEE Transactions on Multimedia},
  title     = {StrongSORT: Make DeepSORT Great Again},
  year      = {2023},
  volume    = {25},
  pages     = {8725--8737},
  doi       = {10.1109/TMM.2023.3240881}
}

@article{afmtrack,
  author    = {Bui, Duy Cuong and Yoo, Myungsik},
  journal   = {IEEE Access},
  title     = {AFMtrack: Attention-Based Feature Matching for Multiple Object Tracking},
  year      = {2024},
  volume    = {12},
  doi       = {10.1109/ACCESS.2024.3411422}
}

@article{pc3t,
  author    = {Wu, Hai and Han, Wenkai and Wen, Chenglu and Li, Xin and Wang, Cheng},
  journal   = {IEEE Transactions on Intelligent Transportation Systems},
  title     = {3D Multi-Object Tracking in Point Clouds Based on Prediction Confidence-Guided Data Association},
  year      = {2022},
  volume    = {23},
  number    = {6},
  pages     = {5668--5677},
  doi       = {10.1109/TITS.2021.3055616}
}

@inproceedings{polarmot,
  author    = {Kim, Aleksandr and Bras{\'{o}}, Guillem and O{\v{s}}ep, Aljo{\v{s}}a and Leal-Taix{\'{e}}, Laura},
  title     = {PolarMOT: How Far Can Geometric Relations Take Us in 3D Multi-Object Tracking?},
  booktitle = {Proceedings of the European Conference on Computer Vision (ECCV)},
  year      = {2022},
  pages     = {41--58},
  publisher = {Springer Nature Switzerland},
  doi       = {10.1007/978-3-031-19772-7_3}
}

@article{eaffmot,
  author    = {Jin, Jingyi and Zhang, Jindong and Zhang, Kunpeng and Wang, Yiming and Pan, Dongyu},
  title     = {3D multi-object tracking with boosting data association and improved trajectory management mechanism},
  journal   = {Signal Processing},
  year      = {2024},
  volume    = {218},
  pages     = {109367},
  doi       = {10.1016/j.sigpro.2023.109367}
}

@article{ug3dmot,
  author    = {He, Jiawei and Wang, Xiyang and Fu, Chunyun and Li, Zhankun and others},
  journal   = {Signal Processing},
  title     = {3D Multi-Object Tracking Based on Informatic Divergence-Guided Data Association},
  year      = {2024},
  volume    = {222},
  pages     = {109521},
  doi       = {10.1016/j.sigpro.2024.109521}
}

@inproceedings{mctrack,
  title={MCTrack: A Unified 3D Multi-Object Tracking Framework for Autonomous Driving},
  author={Wang, Xiyang and Qi, Shouzheng and Zhao, Jieyou and Zhou, Hangning and Zhang, Siyu and Wang, Guoan and Tu, Kai and Guo, Songlin and Zhao, Jianbo and Li, Jian and Yang, Mu},
  booktitle={Proceedings of the IEEE/RSJ International Conference on Intelligent Robots and Systems (IROS)},
  year={2025}
}

@article{robmot,
  author    = {Nagy, Mohamed and Werghi, Naoufel and Hassan, Bilal and Dias, Jorge and Khonji, Majid},
  journal   = {IEEE Transactions on Intelligent Transportation Systems},
  title     = {RobMOT: 3D Multi-Object Tracking Enhancement Through Observational Noise and State Estimation Drift Mitigation in LiDAR Point Clouds},
  year      = {2025},
  volume    = {26},
  number    = {10},
  doi       = {10.1109/TITS.2024.11071990}
}

@misc{nagy,
  title = {Towards Accurate State Estimation: Kalman Filter Incorporating Motion Dynamics for 3D Multi-Object Tracking},
  author = {Nagy, Mohamed and Werghi, Naoufel and Hassan, Bilal and Dias, Jorge and Khonji, Majid},
  year = {2025},
  eprint = {2505.07254},
  archivePrefix = {arXiv},
  primaryClass = {cs.CV},
  url = {https://arxiv.org/abs/2505.07254}
}

@inproceedings{jmodt,
  author    = {Huang, Kemiao and Hao, Qi},
  title     = {Joint Multi-Object Detection and Tracking with Camera-LiDAR Fusion for Autonomous Driving},
  booktitle = {2021 IEEE/RSJ International Conference on Intelligent Robots and Systems (IROS)},
  year      = {2021},
  pages     = {6983--6989},
  doi       = {10.1109/IROS48211.2021.9636006}
}

@article{deepfusion,
  author    = {Wang, Xiyang and Fu, Chunyun and Li, Zhankun and Lai, Ying and He, Jiawei},
  journal   = {IEEE Robotics and Automation Letters},
  title     = {DeepFusionMOT: A 3D Multi-Object Tracking Framework Based on Camera-LiDAR Fusion With Deep Association},
  year      = {2022},
  volume    = {7},
  doi       = {10.1109/LRA.2022.3187264}
}

@article{strongfusionmot,
  author    = {Wang, Xiyang and Fu, Chunyun and He, Jiawei and Wang, Sujuan and Wang, Jianwen},
  journal   = {IEEE Sensors Journal},
  title     = {StrongFusionMOT: A Multi-Object Tracking Method Based on LiDAR-Camera Fusion},
  year      = {2023},
  volume    = {23},
  number    = {1},
  pages     = {472--483},
  doi       = {10.1109/JSEN.2022.3225211}
}

@article{bcmodt,
  author    = {Zhang, Kunpeng and Liu, Yanheng and Mei, Fang and Jin, Jingyi and Wang, Yiming},
  journal   = {Remote Sensing},
  title     = {Boost Correlation Features with 3D-MiIoU-Based Camera-LiDAR Fusion for MODT in Autonomous Driving},
  year      = {2023},
  volume    = {15},
  number    = {4},
  pages     = {874},
  doi       = {10.3390/rs15040874}
}

@article{peng2024pnasmot,
  author    = {Peng, Chensheng and Zeng, Zhaoyu and Gao, Jinling and Zhou, Jundong and Tomizuka, Masayoshi and Wang, Xinbing and Zhou, Chenghu and Ye, Nanyang},
  journal   = {IEEE Robotics and Automation Letters},
  title     = {PNAS-MOT: Multi-Modal Object Tracking With Pareto Neural Architecture Search},
  year      = {2024},
  volume    = {9},
  number    = {5},
  pages     = {4377--4384},
  doi       = {10.1109/LRA.2024.3379865}
}

@InProceedings{tracktrack,
    author    = {Shim, Kyujin and Ko, Kangwook and Yang, Yujin and Kim, Changick},
    title     = {Focusing on Tracks for Online Multi-Object Tracking},
    booktitle = {Proceedings of the IEEE/CVF Conference on Computer Vision and Pattern Recognition (CVPR)},
    month     = {June},
    year      = {2025},
}

@misc{fasttrack,
      title={FastTracker: Real-Time and Accurate Visual Tracking}, 
      author={Hamidreza Hashempoor and Yu Dong Hwang},
      year={2025},
      eprint={2508.14370},
      archivePrefix={arXiv},
      primaryClass={cs.CV},
      url={https://arxiv.org/abs/2508.14370}, 
}

@InProceedings{VirConvDet_CVPR_23,
    author    = {Wu et al.},
    title     = {Virtual Sparse Convolution for Multimodal 3D Object Detection},
    booktitle = {Proceedings of the IEEE/CVF Conference on Computer Vision and Pattern Recognition (CVPR)},
    month     = {June},
    year      = {2023},
}

@ARTICLE{casA_detector,
  author={Wu et al.},
  journal={IEEE Transactions on Geoscience and Remote Sensing}, 
  title={CasA: A Cascade Attention Network for 3-D Object Detection From LiDAR Point Clouds}, 
  year={2022},
  doi={10.1109/TGRS.2022.3203163}}

@InProceedings{point_rcnn_detector,
    author = {Shi, Shaoshuai and Wang, Xiaogang and Li, Hongsheng},
    title = {PointRCNN: 3D Object Proposal Generation and Detection From Point Cloud},
    booktitle = {The IEEE Conference on Computer Vision and Pattern Recognition (CVPR)},
    month = {June},
    year = {2019}
}

@InProceedings{pv-rcnn_detector,
author = {Shi, Shaoshuai and Guo, Chaoxu and Jiang, Li and Wang, Zhe and Shi, Jianping and Wang, Xiaogang and Li, Hongsheng},
title = {PV-RCNN: Point-Voxel Feature Set Abstraction for 3D Object Detection},
booktitle = {Proceedings of the IEEE/CVF Conference on Computer Vision and Pattern Recognition (CVPR)},
month = {June},
year = {2020}
}

@article{second_detector,
  title={Second: Sparsely embedded convolutional detection},
  author={Yan, Yan and Mao, Yuxing and Li, Bo},
  journal={Sensors},
  volume={18},
  number={10},
  pages={3337},
  year={2018},
  publisher={MDPI}
}

@inproceedings{gu2020hippo,
  title     = {HiPPO: Recurrent Memory with Optimal Polynomial Projections},
  author    = {Gu, Albert and Dao, Tri and Ermon, Stefano and Rudra, Atri and R{\'e}, Christopher},
  booktitle = {Advances in Neural Information Processing Systems (NeurIPS)},
  volume    = {33},
  year      = {2020}
}

@article{gu2023mamba,
  title={Mamba: Linear-Time Sequence Modeling with Selective State Spaces},
  author={Gu, Albert and Dao, Tri},
  journal={arXiv preprint arXiv:2312.00752},
  year={2023}
}

@article{kalman1960,
      title={A New Approach to Linear Filtering and Prediction Problems},
      author={Kalman, Rudolph Emil},
      journal={Journal of Basic Engineering},
      volume={82},
      number={1},
      pages={35--45},
      year={1960},
      publisher={American Society of Mechanical Engineers}
    }

@article{mot17,
  title={MOTChallenge: A Benchmark for Single-Camera Multiple Target Tracking},
  author={Dendorfer, Patrick and Osep, Aljosa and Milan, Anton and Schindler, Konrad and Cremers, Daniel and Reid, Ian and Roth, Stefan and Leal-Taix{\'e}, Laura},
  journal={International Journal of Computer Vision},
  volume={129},
  number={4},
  pages={845--881},
  year={2021},
  publisher={Springer},
  doi={10.1007/s11263-020-01393-0}
}

@inproceedings{geiger2012kitti,
  title={Are we ready for Autonomous Driving? The KITTI Vision Benchmark Suite},
  author={Geiger, Andreas and Lenz, Philip and Urtasun, Raquel},
  booktitle={Proceedings of the IEEE Conference on Computer Vision and Pattern Recognition (CVPR)},
  year={2012}
}

@inproceedings{sun2020waymo,
  title={Scalability in Perception for Autonomous Driving: Waymo Open Dataset},
  author={Sun, Pei and Kretzschmar, Henrik and Dotiwalla, Xerxes and Chouard, Aurelien and Gunagi, Vijaysai and Chai, Chengjie and Caine, Benjamin and Vasudevan, Vijay and Han, Wei and Ngiam, Jiquan and others},
  booktitle={Proceedings of the IEEE/CVF Conference on Computer Vision and Pattern Recognition (CVPR)},
  year={2020}
}

@article{hybridsort,
  title={Hybrid-SORT: Weak Cues Matter for Robust Multi-Object Tracking},
  author={Yang, Jinkun and Jiang, Mingzhe and Fan, Zhiyao and others},
  journal={arXiv preprint arXiv:2303.00766},
  year={2023}
}

@inproceedings{bytetrack,
  title={ByteTrack: Multi-object Tracking by Associating Every Detection Box},
  author={Zhang, Yifu and Sun, Peize and Jiang, Yi and Yu, Dongdong and Weng, Fucheng and Yuan, Zehuan and Luo, Ping and Liu, Wenyu and Wang, Xinggang},
  booktitle={Proceedings of the European Conference on Computer Vision (ECCV)},
  year={2022}
}

@software{yolov8_ultralytics,
  author = {Glenn Jocher and Ayush Chaurasia and Jing Qiu},
  title = {Ultralytics YOLOv8},
  version = {8.0.0},
  year = {2023},
  url = {https://github.com/ultralytics/ultralytics},
  orcid = {0000-0001-5950-6979, 0000-0002-7603-6750, 0000-0003-3783-7069},
  license = {AGPL-3.0}
}

@article{eval_1,
  title={HOTA: A Higher Order Metric for Evaluating Multi-Object Tracking},
  author={Luiten, Jonathon and Osep, Aljosa and Dendorfer, Patrick and Torr, Philip and Geiger, Andreas and Leal-Taix{\'e}, Laura and Leibe, Bastian},
  journal={International Journal of Computer Vision},
  pages={1--31},
  year={2020},
  publisher={Springer}
}

@misc{eval_2,
  author =       {Jonathon Luiten, Arne Hoffhues},
  title =        {TrackEval},
  howpublished = {\url{https://github.com/JonathonLuiten/TrackEval}},
  year =         {2020}
}

@INPROCEEDINGS{mobilenet,
  author={Howard, Andrew and Sandler, Mark and Chen, Bo and Wang, Weijun and Chen, Liang-Chieh and Tan, Mingxing and Chu, Grace and Vasudevan, Vijay and Zhu, Yukun and Pang, Ruoming and Adam, Hartwig and Le, Quoc},
  booktitle={2019 IEEE/CVF International Conference on Computer Vision (ICCV)}, 
  title={Searching for MobileNetV3}, 
  year={2019},
  volume={},
  number={},
  pages={1314-1324},
  doi={10.1109/ICCV.2019.00140}}

@article{frigo2005design,
  author = {Frigo, Matteo and Johnson, Steven G.},
  title = {The Design and Implementation of {FFTW3}},
  journal = {Proceedings of the IEEE},
  volume = {93},
  number = {2},
  pages = {216--231},
  year = {2005},
  doi = {10.1109/JPROC.2004.840301}
}

@article{stockham1966high,
  author = {Stockham, Thomas G.},
  title = {High-speed convolution and correlation},
  journal = {AFIPS '66 (Spring): Proceedings of the April 26-28, 1966, Spring Joint Computer Conference},
  pages = {229--233},
  year = {1966},
  doi = {10.1145/1464122.1464147}
}

@article{MOTChallenge20,
  title={MOT20: A benchmark for multi object tracking in crowded scenes},
  author={Dendorfer, P. and Rezatofighi, H. and Milan, A. and Shi, J. and Cremers, D. and Reid, I. and Roth, S. and Schindler, K. and Leal-Taix\'{e}, L.},
  journal={arXiv preprint arXiv:2003.09003},
  year={2020}
}

@INPROCEEDINGS{6252962,
  author={Haynes, David and Corns, Steven and Venayagamoorthy, Ganesh Kumar},
  booktitle={2012 IEEE Congress on Evolutionary Computation}, 
  title={An Exponential Moving Average algorithm}, 
  year={2012},
  volume={},
  number={},
  pages={1-8},
  doi={10.1109/CEC.2012.6252962}}

@ARTICLE{10981717,
  author={Lee, Kyeongtae and Ko, Hankyung and Oh, Donghwan and Kim, Jihye and Oh, Hyunok},
  journal={IEEE Access}, 
  title={Hadamard Product Arguments and Their Applications}, 
  year={2025},
  volume={13},
  number={},
  pages={79736-79756},
  doi={10.1109/ACCESS.2025.3566104}}

@article{yolox2021,
  title={YOLOX: Exceeding YOLO Series in 2021},
  author={Ge, Zheng and Liu, Songtao and Wang, Feng and Li, Zeming and Sun, Jian},
  journal={arXiv preprint arXiv:2107.08430},
  year={2021},
  url={https://arxiv.org}
}

@INPROCEEDINGS{reid,
  author={Luo, Hao and Gu, Youzhi and Liao, Xingyu and Lai, Shenqi and Jiang, Wei},
  booktitle={2019 IEEE/CVF Conference on Computer Vision and Pattern Recognition Workshops (CVPRW)}, 
  title={Bag of Tricks and a Strong Baseline for Deep Person Re-Identification}, 
  year={2019},
  volume={},
  number={},
  pages={1487-1495},
  doi={10.1109/CVPRW.2019.00190}}

@ARTICLE{10771607,
  author={Li, Hui and Yang, Haoran and Ai, Xiaoxue and Chen, Zhong and Wu, Yanli},
  journal={IEEE Transactions on Intelligent Vehicles}, 
  title={A Dynamic 3D Multi-Object Tracking Method Based on Spatiotemporal Features}, 
  year={2025},
  volume={10},
  number={11},
  pages={4974-4991},
  doi={10.1109/TIV.2024.3508743}}

\section{Biography Section}
\begin{IEEEbiography}[{\includegraphics[width=1in,height=1.25in,clip,keepaspectratio]{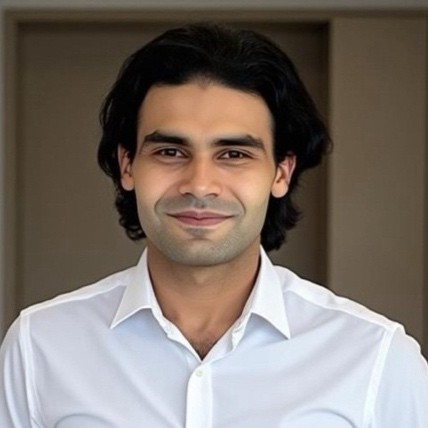}}]{Mohamed Nagy}
received the B.Sc. degree in Mathematics and Computer Science from Helwan University, Egypt, in 2018, and the M.Sc. degree in Computer Science from Khalifa University, UAE, in 2022. He is currently pursuing the Ph.D. degree in Electrical Engineering and Computer Science at Khalifa University, with a focus on autonomous vehicle perception. He is also a visiting researcher at the Technical University of Munich, Munich, Germany. His research interests include object tracking, sensor fusion, and deep learning.
\end{IEEEbiography}

\vspace{11pt}
\begin{IEEEbiography}[{\includegraphics[width=1in,height=1.25in,clip,keepaspectratio]{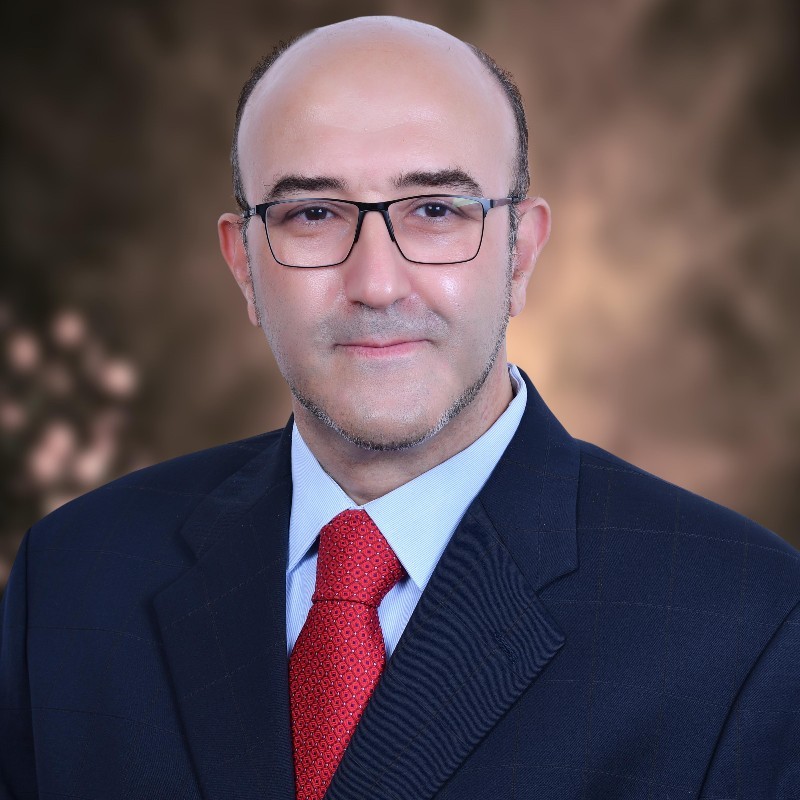}}]{Naoufel Werghi}
received a Research Habilitation and PhD in Computer Vision from the University of Strasbourg. Currently, he is a full Professor in the Department of Computer Sciences and is leading the Artificial Intelligence and Big Data theme at the Center of Secure Cyber-Physical Systems at Khalifa University. His main research area is 2D/3D image/video analysis and understanding, where he has been leading several funded projects related to biometrics, medical imaging robotics, surveillance, inspection, and intelligent systems. He is Associate Editor of the IEEE Transactions on Circuits and Systems for Video Technology, the IEEE Journal of Biomedical and Health Informatics, and the Eurasip  Journal for Image and Video Processing.  
\end{IEEEbiography}

\vspace{11pt}
\begin{IEEEbiography}[{\includegraphics[width=1in,height=1.25in,clip,keepaspectratio]{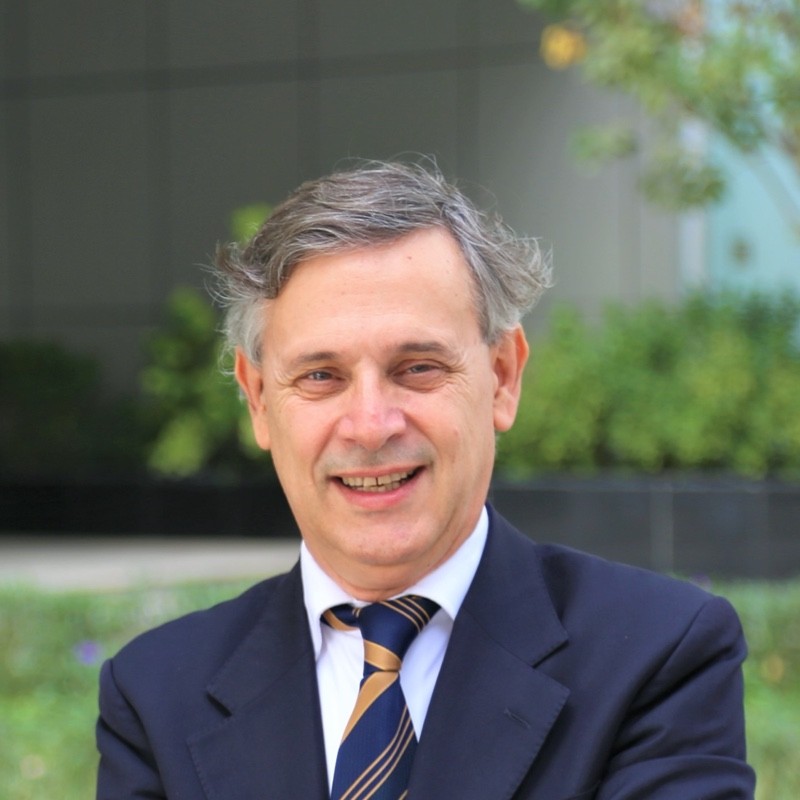}}]{Jorge Dias}
has a Ph.D. in EE and Coordinates the Artificial Perception Group from the Institute of Systems and Robotics from the University of Coimbra, Portugal. He is Full Professor at Khalifa University, Abu Dhabi, UAE and Deputy Director from the Center of Autonomous Robotic Systems from Khalifa University. His expertise is in the area of Artificial Perception (Computer Vision and Robotic Vision) and has contributions on the field since 1984. He has been principal investigator and consortia coordinator from several research international projects, and coordinates the research group on Computer Vision and Artificial Perception from KUCARS. Jorge Dias published several articles in the area of Computer Vision and Robotics that include more than 300 publications in international journals and conference proceedings and recently published book on Probabilistic Robot Perception that addresses the use of statistical modeling and Artificial Intelligence for Perception, Planning and Decision in Robots. He was the Project Coordinator of two European Consortium for the Projects “Social Robot” and “GrowMeUP” that were developed to support the inclusivity and wellbeing for of the Elderly generation. 
\end{IEEEbiography}
\vspace{11pt}
\begin{IEEEbiography}[{\includegraphics[width=1in,height=1.25in,clip,keepaspectratio]{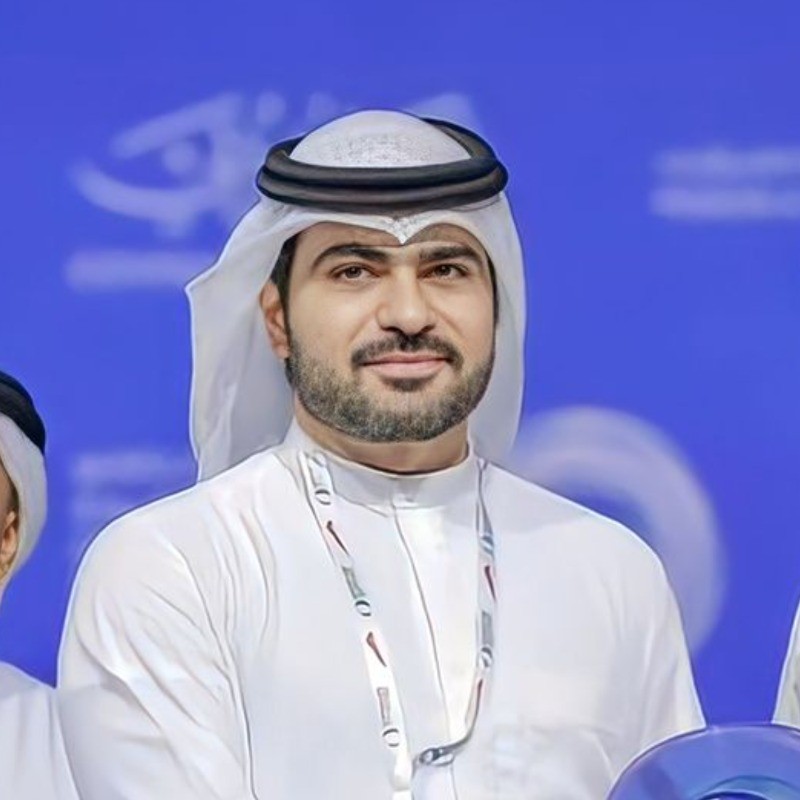}}]{Majid Khonji}
is an Associate Professor in the Computer Science Department at Khalifa University (KU). He leads the research activities in the Autonomous Vehicle Laboratory at the KU Center for Autonomous Robotic Systems (KUCARS). Additionally, he is a research affiliate with the MIT Computer Science and Artificial Intelligence Laboratory (CSAIL) in the USA. Dr. Khonji earned his MSc degree in Security, Cryptology, and Coding of Information Systems from Ensimag, Grenoble Institute of Technology, France, and completed his Ph.D. in Interdisciplinary Engineering at Masdar Institute, in collaboration with MIT, in 2016. Prior to his current role, he was a visiting assistant professor at MIT’s CSAIL, a senior R\&D technologist at Dubai Electricity and Water Authority (DEWA), and an information security researcher at the Emirates Advanced Investment Group (EAIG). His research interests include Probabilistic Planning, Artificial Intelligence, Combinatorial Optimization, Autonomous Vehicles, and Electric Mobility.
\end{IEEEbiography}

\vfill

\end{document}